\documentclass{article}

\usepackage{microtype}
\usepackage{graphicx}
\usepackage{booktabs} % for professional tables

\usepackage{hyperref}

\usepackage[accepted]{icml2024}

\usepackage{amsmath}
\usepackage{amssymb}
\usepackage{mathtools}
\usepackage{amsthm}

\usepackage{multirow}
\usepackage{subcaption}
\usepackage{enumerate}
\usepackage[capitalize,noabbrev]{cleveref}

\theoremstyle{plain}

\theoremstyle{definition}

\theoremstyle{remark}

\icmltitlerunning{Self-Composing Policies for Scalable Continual Reinforcement Learning}

\newcommand{\bm}{\mathbf}

\newcommand{\Actions}{\mathcal{A}}
\newcommand{\States}{\mathcal{S}}
\newcommand{\Rewardf}{r}
\newcommand{\Pdist}{p}
\newcommand{\modi}[1]{#1}

\begin{document}

\twocolumn[
\icmltitle{Self-Composing Policies for Scalable Continual Reinforcement Learning}



\icmlsetsymbol{equal}{*}

\begin{icmlauthorlist}
\icmlauthor{Mikel Malagón}{ehu}
\icmlauthor{Josu Ceberio}{ehu}
\icmlauthor{Jose A. Lozano}{ehu,bcam}
\end{icmlauthorlist}

\icmlaffiliation{ehu}{Department of Computer Science and Artificial Intelligence, University of the Basque Country UPV/EHU, Donostia-San Sebastian, Spain}
\icmlaffiliation{bcam}{Basque Center for Applied Mathematics (BCAM), Bilbao, Spain}

\icmlcorrespondingauthor{Mikel Malagón}{mikel.malagon@ehu.eus}

\icmlkeywords{continual reinforcement learning, reinforcement learning, growable neural networks}

\vskip 0.3in
]



\printAffiliationsAndNotice{}  % leave blank if no need to mention equal contribution

\begin{abstract}
This work introduces a growable and modular neural network architecture that naturally avoids catastrophic forgetting and interference in continual reinforcement learning. The structure of each module allows the selective combination of previous policies along with its internal policy, accelerating the learning process on the current task. Unlike previous growing neural network approaches, we show that the number of parameters of the proposed approach grows linearly with respect to the number of tasks, and does not sacrifice plasticity to scale. Experiments conducted in benchmark continuous control and visual problems reveal that the proposed approach achieves greater knowledge transfer and performance than alternative methods.\footnote{Code available at \url{https://github.com/mikelma/componet}.}

\end{abstract}

\section{Introduction}
\label{sec:introduction}

% The real world is non-stationary. The ability to continuously learn, acquire new knowledge, and fine-tune the existing one, is a crucial stepping stone for Reinforcement Learning (RL) agents that operate in our world \citep{hassabis2017neuroscience}.
The real world is non-stationary. Continuously learning, acquiring new knowledge, and fine-tuning existing skills are vital for Reinforcement Learning (RL) agents operating within our world \citep{hassabis2017neuroscience}.
Dating back to \citet{mnih2013playing}, deep RL has demonstrated the ability to outperform humans in a constantly increasing number of tasks and domains \citep{badia2020agent57,perolat2022mastering}. These powerful algorithms are usually trained \textit{from scratch} to operate in a stationary environment where the goal is to solve a single well-delimited problem. 
On the contrary, humans and other animals greatly benefit from previous experiences to efficiently solve novel tasks \citep{lawrence1952TheTO,elio1984effects}. In this realm, the Continual Reinforcement Learning (CRL) field aims to develop agents that incrementally develop complex behavior based on previously acquired knowledge. Ideally, such agents should learn, adapt, reuse, and transfer knowledge in a never-ending stream of tasks~\citep{hadsell2020embracing}.

It is well known that Neural Networks (NNs) can benefit from the experience obtained in simple problems to approach new and more complex challenges that otherwise would hardly be solvable or would require extreme computational resources \citep{wang2019paired,team2023human}. However, as described by \citet{bengio2009curriculum} and \citet{graves2017automated}, NNs are highly sensitive to the order of appearance and complexity of tasks.
% It is known that deep Neural Networks (NNs) can benefit from the experience obtained in simpler problems to approach new and more complex challenges that otherwise would hardly be solvable or would require extreme computational resources \citep{wang2019paired,team2023human}. However, as described by \citet{bengio2009curriculum} and \citet{graves2017automated},  NNs are highly sensitive to the order in which tasks are presented and their complexity. 
Learning a new task can easily harm the performance of the model in previously learned or future tasks due to the well-known phenomena of \textit{catastrophic forgetting} and \textit{interference} \citep{mccloskey1989catastrophic,french1999catastrophic,kumaran2016learning}.
To overcome the mentioned issues, growable NN architectures \citep{rusu2016progressive,nips2019compacting,gaya2023building} incorporate new NN modules every time a new task is presented. By retaining parameters learned in previous tasks, these methods naturally overcome forgetting and interference, while knowledge is transferred between modules by sharing hidden layer representations. However, by increasing the number of parameters, the memory cost of these models also increases. Indeed, many of these approaches grow quadratically in the number of parameters with respect to the number of tasks, greatly limiting their scalability \citep{terekhov2019knowledge,rusu2016progressive,rusu2017sim}. 

\begin{figure*}[ht]
    \centering
    \includegraphics[width=\textwidth]{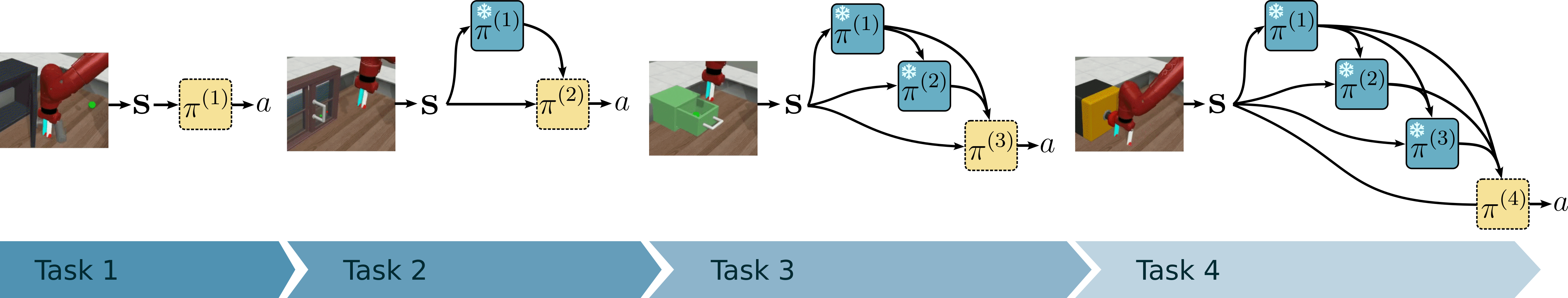}
    % \caption{Evolution of the proposed Self-\textbf{Compo}sing Policies \textbf{Net}work architecture (CompoNet) across multiple tasks.
    % Light yellow blocks indicate trainable policy modules, while dark blue blocks indicate frozen modules. In the first task, a single policy module is trained from scratch. Once the first task ends, the module is frozen and a new trainable module is added. This procedure is repeated in every task-to-task transition. 
    % Note that every module after the first one has access to the current state of the environment together with the outputs of the preceding policy modules.     
    % }
    \caption{Evolution of the Self-\textbf{Compo}sing Policies \textbf{Net}work architecture (CompoNet) across multiple tasks. Trainable self-composing policy modules are represented by light yellow blocks, and frozen modules are denoted by dark blue blocks. The initial task involves training a single policy module from scratch. Following the completion of each task, the trained module is frozen, and a new trainable module is introduced for the subsequent task. This process is repeated in every transition from one task to another. Importantly, each module, after the initial one, benefits from access to the current state of the environment alongside the outputs generated by the preceding policy modules.}
    \label{fig:architecture}
\end{figure*}

In this paper, we present a growable NN architecture that leverages the composition of previously learned policy modules instead of sharing hidden layer representations. This approach significantly reduces the memory and computational cost per task required by the model.
% In this paper, we propose a growable NN architecture that leverages the composition of previously learned policy modules instead of hidden layer representation sharing as the method of knowledge transfer, drastically reducing the number of parameters per task required by the model.
% Contrarily to many compositional NN approaches \citep{rosenbaum2019routing}, the method we introduce does not require the usage of a dedicated NN model that learns to compose the modules, as modules learn to compose themselves (hence the name, \textit{self-composing policies}).
Contrarily to many compositional NN approaches \citep{rosenbaum2019routing}, the method we introduce eliminates the need for a dedicated NN to learn to compose the modules. Instead, modules autonomously learn to compose themselves, hence the name \textit{self-composing policies}.
Illustrated in Figure~\ref{fig:architecture}, the architecture, called CompoNet, adds a new module to the network each time a new task is introduced while retaining modules learned in previous tasks. 
Within the NN architecture of each module, policies from previous modules are selectively composed together with an internal policy, accelerating the learning process for new tasks. Direct access to previously learned modules allows for easy reuse, leverage, and adaptation of stored knowledge. 
This procedure automatically creates a cascading structure of policies that grows in depth with the number of tasks, where each node can access and compose the outputs of previous ones to solve the current task.
%% ORIG
% The approach, named CompoNet, is illustrated in Figure~\ref{fig:architecture}. 
% As depicted, every time a new task is introduced, a new module is added to the network, while retaining the modules learned for solving previous tasks. In turn, the NN architecture of each module can selectively compose the policies from preceding modules together with an internal policy to accelerate the learning process of new tasks. 
% Direct access to previously learned modules enables the one operating in the current task to easily reuse, adapt, and benefit from the knowledge stored in preceding modules.  
% This procedure automatically creates a cascading structure of policies that grows in depth with the number of tasks, where each node can access and compose the outputs of previous ones to solve the current task.

% Performance and Forward transfer
Exhaustive empirical evaluation highlights that CompoNet greatly benefits from the knowledge acquired in solving previous tasks
compared to other CRL methods.
% tasks when facing new problems. 
The latter is demonstrated for sequences of diverse robotic manipulation tasks from the Meta-World environment~\citep{yu2020meta} optimized using the Soft Actor-Critic (SAC) algorithm~\citep{haarnoja2018soft}, as well as in visual control tasks from the Arcade Learning Environment (ALE)~\cite{bellemare13arcade} where Proximal Policy Optimization (PPO)~\citep{schulman2017proximal} is employed.  
% Robustness
% We empirically demonstrate that the presented approach is robust to scenarios where none of the previous policies are informative for solving the current task, being able to learn a policy from scratch with no interference. In contrast, when a previous module can solve the current task, we show that CompoNet effectively reuses it after a few steps of training.
We empirically demonstrate the robustness of the presented approach in scenarios where none of the previous policy modules offer information for solving the current task, enabling the learning of a policy from scratch with no interference. Conversely, when a function over the previous modules solves the current task, we show that CompoNet efficiently learns it after a few steps into the training.
% Scalability
Nevertheless, knowledge transfer and robustness are not the only desirable characteristics of CRL agents: scalability is a critical feature and the potential weak point of growing-size NN approaches \citep{terekhov2019knowledge,rusu2016progressive}. 
Despite the memory costs associated with growth, we demonstrate that the number of parameters required by CompoNet is linear with respect to the number of tasks, greatly enhancing its scalability.
Furthermore, we show that CompoNet efficiently scales in inference time compared to growing NNs from the literature, which demand considerably more time and resources.
Some methods address this challenge by growing only when strictly needed and/or consolidating the knowledge from multiple tasks in a single NN. However, they introduce a dilemma between scalability and the ability to acquire new knowledge (plasticity) \citep{mallya2018packnet,nips2019compacting}. In contrast, the presented approach scales without sacrificing plasticity.

% DINO
%Finally, to further increase the scalability of CompoNet, we show thatrecent pretrained visual foundation models \citep{oquab2023dinov2} can be successfully used to train RL agents, that in the case of CompoNet, allow sharing a visual encoder across modules (see Figure~\ref{fig:architecture}), further reducing the memory cost per task.

%\paragraph{Contributions} We present CompoNet, a modular and growable NN architecture for CRL that naturally avoids catastrophic forgetting and interference. CompoNet scales linearly in the number of parameters with respect to the number of tasks, while not sacrificing plasticity. Experiments on continuous and discrete domain environments show that CompoNet takes advantage of forward knowledge transfer to accelerate learning in streams of tasks.
% As a secondary contribution, we also demonstrate that the features generated by recent visual foundation models can be used to train RL agents without any fine-tuning. 

\section{Related Work}\label{sec:related-work}

The presented work builds upon the extensive literature on CRL, especially on works that explicitly retain knowledge via parameter storage \citep{khetarpal2022continual}.

\textbf{Growable neural networks} increase their capacity every time a new task is encountered, optimizing the newly added parameters to learn the task, while retaining old parameters to avoid forgetting \citep{terekhov2019knowledge,rusu2016progressive,rusu2017sim,czarnecki2018mix}. This procedure allows knowledge transfer from previous NN modules to the current one, naturally avoiding catastrophic forgetting issues at the cost of computational complexity \citep{parisi2018role,hadsell2020embracing}. For example, the number of parameters in \citet{rusu2016progressive} grows quadratically with respect to the number of tasks. 
To address this challenge, recent efforts \citet{yoon2018lifelong,nips2019compacting} prune and selectively retrain parts of old modules, while only adding new modules when needed. However, these methods introduce a trade-off between plasticity and memory cost.

% aim to reduce the computational cost of these types of approaches. 
% In this realm, \citet{yoon2018lifelong} and \citet{nips2019compacting} prune and selectively retrain parts of old modules, while only adding new modules when needed. However, these methods introduce a trade-off between plasticity and memory cost.

\textbf{Neural composition} leverages the composition of specialized NNs to solve complex tasks \citep{rosenbaum2017routing,cases2019recursive,tseng2021robust,khetarpal2022continual,mendez2022reuse}. 
These learning systems have some similarities with the inner workings of the brain; \citet{stocco2010conditional} and \citet{kell2018task} provided evidence of their biological plausibility.
Moreover, recent work has shown that previously learned NNs can be employed to accelerate the learning of new ones \citep{mendez2022modular}, although it requires experience replay to avoid forgetting. However, as described by \citet{rosenbaum2019routing} and \citet{khetarpal2022continual}, these methods require jointly learning the composing strategy and the NNs to compose. This is a non-stationary problem by itself, as the composing strategy depends on the optimization of the NNs being composed and vice versa, making the training process difficult and unstable.

\textbf{Avoiding forgetting by reducing plasticity} has been a promising~\citep{wołczyk2021continual} and popular~\citep{khetarpal2022continual} approach for CRL. This line of research aims to optimize the parameters of an NN in such a way that learning a new task does not interfere with relevant parameters for solving other tasks \citep{kirkpatrick2017overcoming,wortsman2020supermasks,nips2020gradient,nips2021conflict}. 
For example, \citet{kirkpatrick2017overcoming} propose to selectively slow down the learning of parameters relevant to other tasks. Alternatively, \citet{mallya2018packnet} and \citet{wortsman2020supermasks} decompose the parameters into subnetworks that can be retrieved according to the current task.
Although these methods effectively overcome forgetting in NNs, this ability comes with the cost of limited plasticity, bringing about the \textit{stability-plasticity dilemma} \citep{mermillod2013stability,khetarpal2022continual}.

\section{Preliminaries} \label{sec:preliminaries}

We start by introducing the notation employed in the rest of the paper and formalizing the problem to solve. 

%% ==> Brief RL classic definition + what do we consider a TASK
The classical RL environment \citep{sutton2018reinforcement} is defined as a Markov Decision Process (MDP) in the form of $M = \langle \States, \Actions, \Pdist, \Rewardf, \gamma \rangle$, where $\States$ is the space of states, $\Actions$ is the space of actions,  
$\Pdist: \States \times \States \times \Actions \rightarrow [0, 1]$ is the state transition probability function,
$\Rewardf: \States \times \Actions \rightarrow \mathbb{R}$ is the reward function, and $\gamma \in [0, 1]$ is the discount rate. At every timestep,
the agent receives a state $\bm{s}\in\States$ of the environment, and takes an action $a\in\Actions$ sampled from the probability function $\pi: \Actions \times \States \rightarrow [0, 1]$, known as the policy.  
Then, a new state $\bm{s}'$ is sampled $\bm{s'} \sim \Pdist(\cdot|\bm{s}, a)$ and
the reward $\Rewardf$ is computed.
The objective is to find the optimal policy $\pi^*$ that maximizes the expected sum of discounted rewards for all states in $\States$.

%%% ====> Definition of our problem
Aligned with the CRL definition by \citet{khetarpal2022continual}, 
we characterize non-stationary environments as MDPs whose components might exhibit some time dependence. 
We define a task $k$ as a stationary MDP $M^{(k)} = \langle \States^{(k)}, \Actions^{(k)}, \Pdist^{(k)}, \Rewardf^{(k)}, \gamma^{(k)} \rangle$, where $k$ is discrete and changes over time, creating a sequence of tasks. Specifically, we consider that the agent has some limited budget of timesteps $\Delta^{(k)}$ to interact with a task $M^{(k)}$, to optimize the policy $\pi^{(k)}$. After consuming the budget, a new task $M^{(k+1)}$ is introduced, and the agent is limited to only interacting with this task.
The aim is to accelerate and enhance the optimization of the policy $\pi^{(k)}$ to solve the task $M^{(k)}$ leveraging the knowledge of the previous policies $\{\pi^{(i)}\}_{i=1,\ldots,k-1}$.

%%% ===> Assumptions
\textbf{Assumptions.} We take into consideration three usual assumptions on the types of variations that might occur between different tasks~\citep{wołczyk2021continual,khetarpal2022continual}.
First, the action space $\Actions$ remains constant across all tasks. We consider this assumption soft, as tasks with different groups of actions can be considered to share a
single set of actions $\Actions$ by setting the probability of sampling the other actions to zero depending on the task. 
Secondly, task transition boundaries and their identifiers are known to the agent, as usually assumed in the literature~\citep{wołczyk2021continual,wolczyk2022disentangling,khetarpal2022continual}.
Finally, the variations between tasks mostly occur in the \textit{underlying logic} of the tasks, mainly determined by $\Pdist$ and $\Rewardf$. Consequently, tasks should have a similar state space, $\States^{(i)} \approx \States^{(j)}$, as generalization across different state spaces is mostly related to the domain adaptation problem and invariant representation learning literature \citep{higgins2017darla,zhang2020learning}.

% For this work, given a sequence of previously learned policies $\{\pi^{(i)}\}_{i=1:k-1}$, where the policy $\pi^{(i)}$ faced task $M^{(i)}$,
% we recognize three scenarios regarding the relationship between the current task $M^{(k)}$ and the set of previous policies: 
% (1) $M^{(k)}$ can be directly solved by a previously learned policy; 
% (2) $M^{(k)}$ can be directly solved by a function involving the previous policies and the current state; 
% (3) the current task can neither be directly solved by a previous policy nor a by function over the previous policies and the current state. 
% Note that the first case is a particular scenario of the second one, where the function always returns the policy that solves the task.
% The design of CompoNet, presented in the next section, is strongly oriented to handle and exploit the described relations.

For this work, given a sequence of previously learned policies $\{\pi^{(i)}\}_{i=1,\ldots,k-1}$, where the policy $\pi^{(i)}$ corresponds to task $M^{(i)}$, we identify three scenarios regarding the relationship between the current task $M^{(k)}$ and the set of previous policies: (i) $M^{(k)}$ can be directly solved\footnote{\label{note:solve}
%
% By \textit{solve} we refer to achieving relatively high reward or success in the task at hand, not necessarily being the optimal policy.
By \textit{solve} we refer to achieving a sum of rewards in an episode (e.g., time until game over) above a certain threshold defined by the task at hand, not necessarily being the optimal policy.
} by a previously learned policy; (ii) $M^{(k)}$ can be solved by a function involving the previous policies and the current state; (iii) the current task cannot be solved by either a previous policy or a function based on the previous policies. Note that the first case is a specific instance of the second, where the function always returns the policy solving the task; we distinguish it as an especially relevant case. The design of CompoNet, presented in the next section, is specifically tailored to handle and exploit the mentioned scenarios.

\section{Model Architecture} \label{sec:model}

Concerning the three scenarios outlined in the preceding paragraph,
our goal is to design a learning system that fulfills the following desiderata: in scenario (i) the model should exploit the previous policy that solves the current task; in (ii) the model should learn the specified function; in (iii) the agent should learn the policy that solves the task from scratch with minimal interference from previous policies. 

The basic unit of the CompoNet architecture is the self-composing policy module, depicted as blocks in Figure~\ref{fig:architecture}. Whenever a new task $M^{(k)}$ is presented, the parameters of the policy from the preceding task $\pi^{(k-1)}$ are frozen, and a new learnable module corresponding to $\pi^{(k)}$ is introduced. %While operating in the $k$-th task, states $\bm{s}\in\States^{(k)}$ are provided to all the previous modules, obtaining a set of $k-1$ probability distributions over the action space $\Actions$.
While operating in the $k$-th task, states $\bm{s}\in\States^{(k)}$ are provided to all the previous modules, obtaining a set of $k-1$ output vectors, one for each module $j\in\{1,\ldots,k-1\}$.
For the sake of compact notation, we will refer to the mentioned set of outputs as the matrix $\Phi^{k;\bm{s}}$ with dimensions $(k-1) \times |\Actions|$, where the $j$-th row $\Phi^{k;\bm{s}}_j$ 
%corresponds to the probability vector of $\pi^{(i)}$ when $\Actions$ is discrete, and to the mean 
corresponds to the output vector of the $j$-th module. Note that these vectors define the probability values of a categorical distribution over $\Actions$ when actions are discrete, and to the mean vector of a Gaussian distribution when actions are continuous.
Then, we can recursively define any policy $\pi^{(k)}$ of the architecture as the probability distribution over $\Actions$ conditioned on the current state $\bm{s}\in\States^{(k)}$ and the matrix $\Phi^{k;\bm{s}}$, formally, $\pi^{(k)}(a|\bm{s}, \Phi^{k;\bm{s}})$.

Therefore, every time a new task is presented to the agent, its NN architecture is changed by adding a new policy module to the cascading graph structure (see Figure~\ref{fig:architecture}), increasing in depth by a unit.   
In turn, the new module is not limited to accessing only the current state $\bm{s}$, but it can also benefit from the policies of preceding modules. 
%This allows the model to exploit the relations described in Section~\ref{sec:preliminaries}, between the learned policies and the current task.  
This allows the model to exploit the relations between the learned policies and the current task described in the last part of Section~\ref{sec:preliminaries}.

\subsection{State Encoding}\label{sec:encoder}

To ensure the effectiveness of CompoNet, the input of each module must remain consistent across tasks: the input state distribution of a module cannot change once its parameters are frozen. In this section, we contemplate different state encoding strategies depending on the nature of the tasks.
% RL agents make decisions (at least) based on the current state, this section discusses several options for encoding state representation within the CompoNet architecture. 

When states have a large dimensional representation (e.g., multiple RGB images), we consider an encoder for each module. Usually, these encoders might be defined by simple CNNs that reduce states $\bm{s}\in\States$ into a lower dimensional space $\bm{h_s}\in\mathbb{R}^{d_\text{enc}}$ where the most useful features are retained. Note that this approach is only viable when states are low-resolution images requiring simple CNNs; as is the case for many RL benchmarks such as the arcade learning environments~\citep{bellemare13arcade}.

% When states have a large dimensional representation (e.g. multiple RGB images), 
% it is a common practice to encode states $\bm{s}\in\States$ in a lower dimensional latent space $\bm{h_s}\in\mathbb{R}^{d_\text{enc}}$, where the most useful features are retained. 
% In the CompoNet architecture, where every module is partially conditioned by the current state (see Figure~\ref{fig:architecture}), instantiating an encoder for each module is only feasible when the encoder consists of a relatively small NN such as simple CNN. Note that this is the case for many RL benchmarks employed in the literature, where states encompass very low-resolution images, as in the arcade learning environments~\citep{huang2022cleanrl}.

Although less common, when states comprise high-dimensional images, or when dealing with an extreme number of tasks, a single encoder shared across all tasks can be used. 
%Note that to ensure the effectiveness of CompoNet, the state representation must remain consistent across tasks, as the distribution of the input space for a module cannot change once its parameters are frozen.
For example, leveraging recent vision foundational models for generating representations without requiring fine-tuning or prior task-specific information~\citep{oquab2023dinov2}.\footnote{Refer to Appendix~\ref{sec:dino-appendix} for preliminary results analyzing the feasibility of this approach.}

Otherwise, when states are low dimensional real-valued vectors no encoder is needed for CompoNet, thus $\bm{h_s} = \bm{s}$. 

\subsection{Self-Composing Policy Module} \label{sec:self-compo-pol}

\begin{figure*}[t]
    \centering
    \includegraphics[width=0.9\textwidth]{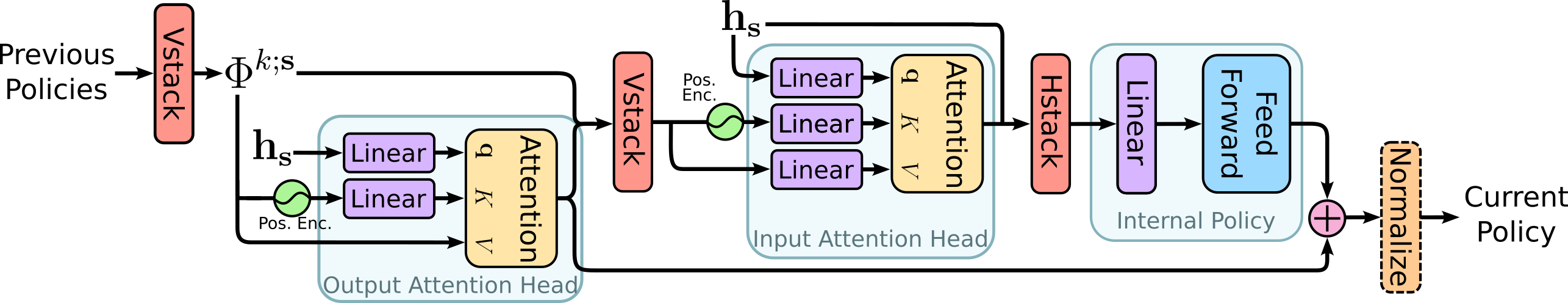}
    \caption{Diagram of the self-composing policy module. \textit{Vstack} and \textit{Hstack} successively represent row-wise and column-wise concatenation operations, while the normalization operation has been delimited with a dashed line to denote that it is optional and dependent on the nature of the action space. Finally, note that the only blocks with learnable parameters are the feed-forward block and the linear transformations.}
    \label{fig:compo-unit}
\end{figure*}

In this section, the building block of the presented architecture is described: the \textit{self-composing policy module}. 
As illustrated in Figure~\ref{fig:compo-unit}, it is divided into three main blocks; corresponding to the light-blue areas of the diagram: the output attention head, the input attention head, and the internal policy. 
% In the following lines, the three blocks are individually described.
The following lines describe and explain the rationale behind these three blocks.  

\paragraph{Output Attention Head.} 
% The output attention head is the first block of the self-composing policy module. 
% Intuitively, the block is used to propose an output for the module directly based on the policies of the preceding modules. 
This block proposes an output for the current module directly based on the preceding policies. 
Specifically, it generates a tentative vector $\bm{v}$ for the output of the module based on the matrix $\Phi^{k;\bm{s}}$ and the representation of the current state $\bm{h_s}$.
In fact, the block employs an attention mechanism that, conditioned on $\bm{h_s}$, returns a linear combination of the outputs of the previous policies (the rows of $\Phi^{k;\bm{s}}$). 
The query vector is obtained as the linear transformation $\bm{q} = \bm{h_s}W_\text{out}^Q$, where $W_\text{out}^Q \in \mathbb{R}^{d_{\text{enc}} \times d_\text{model}}$ is a parameter matrix and $d_\text{model}$ is the hidden vector size. The keys matrix is computed as $K = (\Phi^{k;\bm{s}} + E_\text{out}) W_\text{out}^K$, where $W_\text{out}^K\in\mathbb{R}^{|\Actions| \times d_\text{model}}$ is a parameter matrix, and $E_\text{out}$ is a positional encoding matrix of the same size of $\Phi^{k;\bm{s}}$. The positional encoding method considered in this work is the cosine positional encoding of \citet{vaswani2017attention}. In the case of the values matrix, no linear transformation is considered, thus, $V = \Phi^{k;\bm{s}}$. The result of the output attention head is the scaled dot-product attention~\citep{vaswani2017attention} of $\bm{q}$, $K$, and $V$:  
\begin{equation} \label{eq:attention}
    \text{Attention}(\bm{q}, K, V) = \text{softmax}\left( \frac{qK^T}{\sqrt{d_\text{model}}} \right) V
\end{equation}

%Specifically, the block employs an attention mechanism that, depending on $\bm{h_s}$, composes the outputs of all previous modules together with the output of the internal policy, resulting in the final probability distribution outputted by the self-composing policy module.  
%Unlike the input attention head, in this case, the attention operation is performed directly over the (row-wise) concatenation of $\Phi^{k;\bm{s}}$ with the output of the internal policy, referred to as the matrix $E$. No linear projection is performed to obtain the values $V$, thus $V = E$. The query and keys are computed as $\bm{q} = \bm{h_s}W_{out}^Q$ and $K = (M + P) W_{out}^K$, where $W_{out}^Q\in\mathbb{R}^{d_\text{enc} \times d_\text{model}}$ and $W_{out}^K \in \mathbb{R}^{|\Actions| \times d_\text{model}}$, and $P$ is the positional encoding. Once the query, key, and values are calculated, the result of the output attention head is computed following the same scaled dot-product attention as in Equation~\eqref{eq:attention}. 

\paragraph{Input Attention Head.} 
% The purpose of this block is to extract useful information from the previous modules and the output attention head for the internal policy (the next block). 
The purpose of this block is to retrieve relevant information from both the previous policies and the output attention head. It provides the necessary information for the decision-making process of the internal policy (the next block) by attending to the important features from past policies and the tentative vector $\bm{v}$ from the output attention head.
Similarly to the previous block, it employs an attention head conditioned on $\bm{h_s}$, but unlike the preceding block, the attention head returns a linear combination over learnable transformations of its inputs.
Specifically, the query vector is computed as $\bm{q} = \bm{h_s}W_\text{in}^Q$, where $W_\text{in}^Q \in \mathbb{R}^{d_{\text{enc}} \times d_\text{model}}$. Following Figure~\ref{fig:compo-unit}, the keys are computed as $(P + E_\text{in})W_\text{in}^K$, where $P$ is the row-wise concatenation of the output of the previous block ($\bm{v}$) and $\Phi^{k;\bm{s}}$, while $E_\text{in}$ is a positional encoding matrix of the same size as $P$ and $W_\text{in}^K\in\mathbb{R}^{|\Actions|\times d_\text{model}}$. In turn, the values matrix is obtained as the linear transformation $V = P W_\text{in}^V$, where $W_\text{in}^V\in\mathbb{R}^{|\Actions|\times d_\text{model}}$. Once $\bm{q}$, $K$, and $V$ have been computed, the output of this block is the dot-product attention of these three elements, see Equation~\eqref{eq:attention}. Note that the learnable parameters of this block are $W_\text{in}^Q$, $W_\text{in}^K$, and $W_\text{in}^V$.

\paragraph{Internal Policy.} 
This block is used to adjust, overwrite, or retain the tentative vector $\bm{v}$ from the output attention head, considering the contextual information provided by the input attention head and the representation of the current state.
It is comprised of a feed-forward multi-layer perceptron network which takes the result of the previous block and $\bm{h_s}$ as input, generating a real-valued vector of size $|\Actions|$.
Notably, this vector is not the direct output of the self-composing policy module; instead, it is added to the tentative vector $\bm{v}$ to form the final output of the module. 
Finally, depending on the nature of the task at hand, this addition might require normalization,  
as the output of the module usually represents a categorical distribution over $\Actions$ or continuous actions within some bounds. 

In summary, the output attention head proposes an output for the module based on the current state and the information of the previous policies. 
Subsequently, the input attention head retrieves relevant information from both the previous policies and the output attention head. 
Then, the internal policy utilizes this information along with the current state representation to adjust, overwrite, or retain the tentative output from the output attention head.
Note that the output attention head proposes outputs based on preceding policies and the current state representation, while the input attention head retrieves relevant information from previous modules for guiding the decision-making process of the internal policy.

%Under the proposed setting, the only learnable parameters of CompoNet are 
%parameters of the last module are learned in each task maintaining the previous modules frozen. Moreover, the only learnable parameters of the module are the linear transformations of the attention heads and the parameters of the internal policy.
 
%In Section~\ref{sec:preliminaries} we categorized three possible scenarios regarding the current task and previously learned policies that served to form a desiderata that has driven the design of the proposed architecture.  The following lines review these objectives  under the described architecture:

In Section~\ref{sec:preliminaries}, we categorized three scenarios concerning the current task and previously learned policy modules that motivated the design of CompoNet. The subsequent lines review these scenarios within the described architecture:
\vspace{-2mm}
\begin{enumerate}[(i)]  
    \setlength\itemsep{0.1em}
    % \item If there is a previous module that solves the current task, the output attention head can assign high attention to it, while the internal policy can output a vector of zeros to leave the result of the output attention head intact. This final behavior resembles the residual connections commonly used in the deep learning literature~\citep{he2016residual}. 
    %\item If a previous module solves the current task, the output attention head may assign it high attention, while the internal policy can output a vector of zeros to retain the result of the output attention head. This behavior resembles the residual connections often employed in deep learning literature.
    \item If a previous policy solves the current task, the output attention head assigns high attention to it, and the internal policy may output a vector of zeros to retain this result, akin to residual connections in deep learning.
    \item If a function over the previous policies and the current state can solve the task at hand, then the three blocks of the module can be used to learn such a function.
    % \item In case there is no informative previous policy for solving the current task, the internal policy can learn to solve the current task from scratch, as it has direct access to the information of the current state. In this case, the internal policy would overwrite the result of the output attention head in the final addition and normalization steps (see Figure~\ref{fig:compo-unit}).
    % \item When no previous policy offers relevant information to solve the current task, the internal policy can independently learn to address the task using its direct access to the current state information. In such cases, the internal policy would supersede the result of the output attention head during the final addition and normalization steps (see Figure~\ref{fig:compo-unit}).
    \item When previous policies offer no relevant information for the task at hand, the internal policy independently learns a policy from scratch based on current state information, superseding the result of the output attention head in the last addition step.
\end{enumerate} 

\subsection{Computational Cost}

\begin{figure}[ht]
    \vskip 0.1in
    \centering  
    \includegraphics[width=0.45\textwidth]{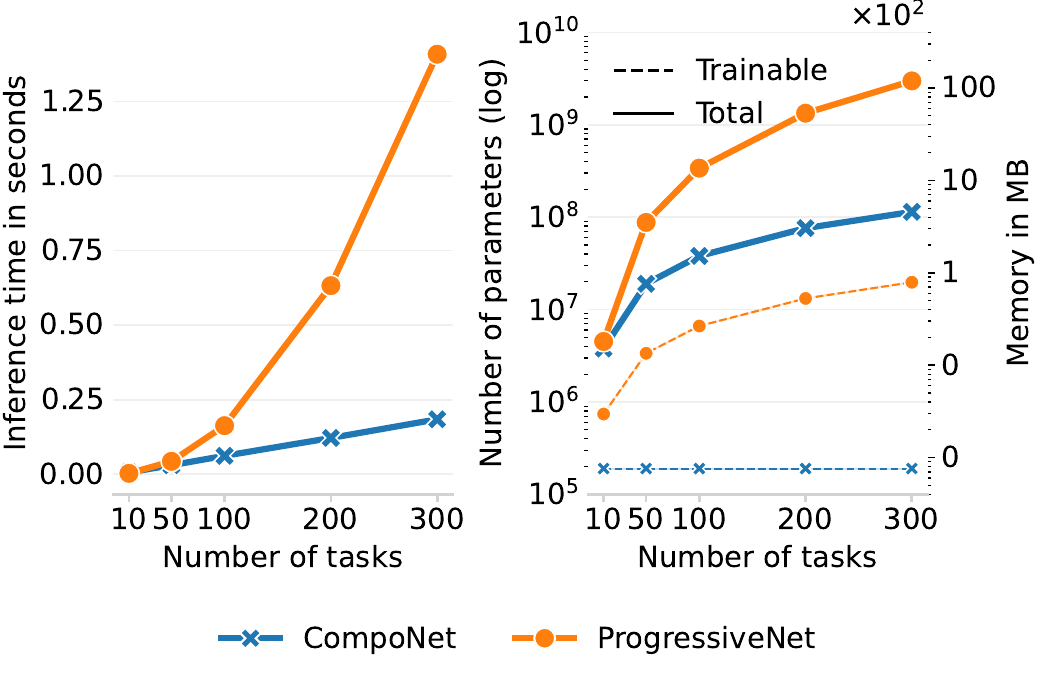}    
     \caption{
     Empirical computational cost of inference (left) and growth in the number of parameters (right) with respect to the number of tasks for CompoNet and ProgressiveNet methods. Hyperparameters are: $d_\text{enc}=64$, $|\mathcal{A}|=6$, $d_\text{model}=256$, and a batch size of 8. Measurements have been taken in a machine with an AMD EPYC 7252 CPU and an NVIDIA A5000 GPU.}
     %The hyperparameters were set according to the settings used in the experimentation section: dimension of the observation vector was 64, $|\mathcal{A}|=6$, and $d_\text{model}=256$. The measurements were run in a machine with an AMD EPYC 7252 CPU and a NVIDIA A5000 GPU.
     %The leftmost figure indicates the empirical computational cost of inference with respect to the number of tasks for both CompoNet and ProgressiveNet methods. The rightmost figure shows the growth in the number of total and trainable parameters of these approaches as the number of tasks increase. The hyperparameters where set according to the settings used in the experimentation section: dimension of the observation vector was 64, $|\mathcal{A}|=6$, and $d_\text{model}=256$. The measurements were run in a machine with an AMD EPYC 7252 CPU and a NVIDIA A5000 GPU.
     \label{fig:benchmarking}
     \vskip -0.1in
\end{figure}

As mentioned in Section~\ref{sec:related-work}, computational cost stands as the primary drawback of growing NN architectures. 
% Consequently, memory and inference costs have been a major concern for this work. 
Consequently, memory and inference costs are focal points in this study.
First, the self-composing policy module is designed to mitigate the memory complexity of the model.
As a result, CompoNet grows linearly in the number of parameters with respect to the number of tasks while being able to encompass the information of all previously learned modules (see Appendix~\ref{sec:memory-cost-appendix} for further details). The rightmost plot in Figure~\ref{fig:benchmarking} contrasts the memory costs of CompoNet with progressive NNs (ProgressiveNet). The latter method, introduced by \citet{rusu2016progressive} is one of the best-known growing NNs, and shares multiple similarities with CompoNet.
Regarding the computational cost of inference, while the theoretical cost of CompoNet is quadratic with respect to the number of tasks (elaborated in Appendix~\ref{sec:complexity-inference}), the results presented in Figure~\ref{fig:benchmarking} indicate that the empirical computational cost of CompoNet does not exhibit quadratic growth up to the 300 tasks tested, effectively scaling to very long task sequences in practice.

% Although this cost is far from ideal, we argue that the real-life inference cost of the model is minimal compared to previous growing NN approaches such as ProgressiveNet under the same parameters, and that effectively scales to very long task sequences in the practice, see Figure~\ref{fig:benchmarking}.
%
% Regarding the computational cost of inference, CompoNet has a quadratic cost with respect to the number of tasks (see Appendix~\ref{sec:complexity-inference}). Although this cost is far from ideal, we argue that the real-life inference cost of the model is minimal compared to previous growing NN approaches such as ProgressiveNet under the same parameters, and that effectively scales to very long task sequences in the practice, see Figure~\ref{fig:benchmarking}.

\section{Experiments} \label{sec:experiments}

In this section, we validate the presented architecture across sequences of tasks from multiple environments and domains. The central hypothesis of these experiments is that \mbox{CompoNet} should be able to benefit from forward knowledge transfer to solve the task at hand 
in the scenarios presented in Section~\ref{sec:model}.
%by fulfilling the desiderata presented in the first lines of Section~\ref{sec:model}.

\subsection{Evaluation Metrics}\label{sec:metrics}

We start by describing the CRL-relevant metrics commonly used in the literature \citep{wołczyk2021continual,wolczyk2022disentangling}.  
Consider $p_i(t)\in [0, 1]$ to be the success rate\footnote{We use $p_i(t)$ to denote success rate (performance) as it is the standard notation in the literature. Not to be confused with $p$, which is commonly used for probability functions.} in task $i$ at time $t$, indicating whether the task is solved\textsuperscript{\ref{note:solve}}, $p_i(t)=1$, or not, $p_i(t)=0$. Note that the metric is task-specific and defined by the problem itself.
Moreover, the interaction of an agent with each task is limited to $\Delta$ timesteps, being the total number of timesteps $T = N \cdot \Delta$, where $N$ is the number of tasks. Continuing the standard practice in CRL, we consider an agent trained from scratch in each task as the baseline for the following metrics~\citep{diaz2018don,wołczyk2021continual,wolczyk2022disentangling}.  

\paragraph{Average Performance.} The average performance at timestep $t$ is computed as ${\text{P}(t) = \frac{1}{N} \sum_{i=1}^N p_i(t)}$. In the next sections, we report the final performance value $\text{P}(T)$ as it is a commonly used metric in the CL literature~\citep{wołczyk2021continual}.

\paragraph{Forward Transfer.} The forward transfer is defined as the normalized area between the training curve of the method and the training curve of the baseline. Considering $p_i^b(t)\in[0,1]$ to be the performance of the baseline, the forward transfer $\text{FTr}_i$ on task $i$ is,
\begin{equation}
\begin{gathered}
    \text{FTr}_i = \frac{\text{AUC}_i-\text{AUC}_i^b}{1-\text{AUC}_i^b}, \quad \text{AUC}_i = \frac{1}{\Delta} \int_{(i-1)\cdot\Delta}^{i\cdot\Delta} p_i(t) \mathrm{d}t , \\
    \text{AUC}_i^b = \frac{1}{\Delta} \int_0^{\Delta} p_i^b(t) \mathrm{d}t
\end{gathered}
\label{eq:ft}
\end{equation}
%Under this metric, an important concept is the Reference forward Transfer (RT). A good CRL method is expected to perform at least as good as fine-tuning from the task of highest transfer to the current one. Thus, RT is defined as the following as,
Under this metric, a key concept is the \textbf{Reference forward Transfer (RT)}. A CRL method should ideally perform at least as well as fine-tuning from the task with the highest transfer to the current one. Thus, RT is defined as follows:
\begin{equation}
    \text{RT} = \frac{1}{N} \sum_{i=2}^N \max_{j<i} \text{FTr}(j, i)
\label{eq:rt}
\end{equation}
where $\text{FTr}(j, i)$ is the forward transfer obtained by training a model from scratch in the $j$-th task and fine-tuning it in the $i$-th task. Note that a model can outperform the RT by composing the knowledge from previous tasks~\citep{wołczyk2021continual}. 

% With respect to the forgetting metric also employed in the CRL literature, we define a provide results for this metric in Appendix~\ref{sec:further-results}, as CompoNet naturally avoids forgetting under the employed assumptions.

% Regarding the forgetting metric used in CRL literature, we define and present results for this metric in Appendix~\ref{sec:further-results}, as CompoNet naturally mitigates forgetting under the employed assumptions.

% \paragraph{Forgetting.} 
% For a task $i$, forgetting is measured as ${\text{F}_i = p_i(i\cdot\Delta) - p_i(T)}$. Note that in the case where task identifiers are known to the agent (as it is in this paper, see the assumptions from Section~\ref{sec:preliminaries}), CompoNet naturally avoids forgetting: if an already solved task is revisited, the policy module that was learned in that task could be used. 

\begin{table*}[ht]
\caption{
% Aggregated results from all of the considered task sequences and methods. The average and standard deviation of 10 random seeds are given for the metrics described in Section~\ref{sec:metrics} and best results are highlighted in bold. The last row corresponds to the reference forward transfer for each sequence. CompoNet, the method proposed in this paper, obtains the highest performance and forward transfer in all three sequences.
Summary of results across all task sequences and methods. Metrics from Section~\ref{sec:metrics} are presented as averages and standard deviations from 10 random seeds, with the best results highlighted in bold. The last row indicates the reference forward transfer (RT) for each sequence. CompoNet, the method proposed in this paper, achieves superior performance and forward transfer in all three sequences.
}
\label{table:main-results}
%\vskip 0.15in
\vskip 0.1in
\begin{center}
\begin{small}
\begin{sc}
\scalebox{0.9}{
\begin{tabular}{lccccccccc}
\toprule 
 & \multicolumn{2}{c}{Meta-World} & & \multicolumn{2}{c}{SpaceInvaders} & & \multicolumn{2}{c}{Freeway} \\ 
\cmidrule{2-3} \cmidrule{5-6} \cmidrule{8-9} 
\multicolumn{1}{l}{Method} & Perf. & \multicolumn{1}{c}{Fwd. Transf.} & & Perf. & \multicolumn{1}{c}{Fwd. Transf.} & & Perf. & \multicolumn{1}{c}{Fwd. Transf.} \\ \midrule
% From scratch
\multicolumn{1}{l|}{Baseline}
& 0.06\scalebox{0.7}{$\pm 0.12$} & \multicolumn{1}{c}{0.00\scalebox{0.7}{$\pm 0.00$}} % Meta-World
& & 0.56\scalebox{0.7}{$\pm 0.37$} & \multicolumn{1}{c}{0.00\scalebox{0.7}{$\pm 0.00$}} % SpaceInvaders
& & 0.19\scalebox{0.7}{$\pm 0.26$} &  \multicolumn{1}{c}{0.00\scalebox{0.7}{$\pm 0.00$}} % Freeway
\\
% Finetune:
\multicolumn{1}{l|}{FT-1}
& 0.03\scalebox{0.7}{$\pm 0.09$} & \multicolumn{1}{c}{-0.21\scalebox{0.7}{$\pm 0.38$}}  % Meta-World
& & 0.44\scalebox{0.7}{$\pm 0.50$} & \multicolumn{1}{c}{0.73\scalebox{0.7}{$\pm 0.25$}} % SpaceInvaders
& & 0.15\scalebox{0.7}{$\pm 0.36$} & \multicolumn{1}{c}{0.64\scalebox{0.7}{$\pm 0.09$}} % Freeway
\\
\multicolumn{1}{l|}{FT-N}
& 0.37\scalebox{0.7}{$\pm 0.48$} & \multicolumn{1}{c}{-0.21\scalebox{0.7}{$\pm 0.38$}}  % Meta-World
& & \textbf{0.99}\scalebox{0.7}{$\bm{\pm 0.01}$} & \multicolumn{1}{c}{0.73\scalebox{0.7}{$\pm 0.25$}} % SpaceInvaders
& & 0.81\scalebox{0.7}{$\pm 0.01$} & \multicolumn{1}{c}{0.64\scalebox{0.7}{$\pm 0.09$}} % Freeway
\\
% ProgressiveNet:
\multicolumn{1}{l|}{ProgNet}
& 0.41\scalebox{0.7}{$\pm 0.49$} & \multicolumn{1}{c}{-0.04\scalebox{0.7}{$\pm 0.04$}} % Meta-World
& & 0.71\scalebox{0.7}{$\pm 0.25$} & \multicolumn{1}{c}{0.10\scalebox{0.7}{$\pm 0.07$}} % SpaceInvaders
& & 0.47\scalebox{0.7}{$\pm 0.28$} & \multicolumn{1}{c}{0.30\scalebox{0.7}{$\pm 0.18$}} % Freeway
\\
% PackNet:
\multicolumn{1}{l|}{PackNet}
& 0.24\scalebox{0.7}{$\pm 0.40$}  & \multicolumn{1}{c}{-0.67\scalebox{0.7}{$\pm 1.38$}} % Meta-World
& & 0.63\scalebox{0.7}{$\pm 0.33$} & \multicolumn{1}{c}{0.36\scalebox{0.7}{$\pm 0.31$}} % SpaceInvaders
& & 0.51\scalebox{0.7}{$\pm 0.20$} & \multicolumn{1}{c}{0.31\scalebox{0.7}{$\pm 0.25$}} % Freeway
\\
% CompoNet:
%\midrule
\multicolumn{1}{l|}{\textbf{CompoNet}}
& \textbf{0.42}\scalebox{0.7}{$\bm{\pm 0.49}$} & \multicolumn{1}{c}{\textbf{0.01}\scalebox{0.7}{$\bm{\pm 0.14}$}} % Meta-World
& & \textbf{0.99}\scalebox{0.7}{$\bm{\pm 0.01}$} & \multicolumn{1}{c}{\textbf{0.74}\scalebox{0.7}{$\bm{\pm 0.22}$}} % SpaceInvaders
& & \textbf{0.94}\scalebox{0.7}{$\bm{\pm 0.06}$} & \multicolumn{1}{c}{\textbf{0.80}\scalebox{0.7}{$\bm{\pm 0.07}$}} % Freeway
\\
\midrule
\multicolumn{1}{l|}{RT}
& --- & \multicolumn{1}{c}{-0.06} % Meta-World
& & --- & \multicolumn{1}{c}{0.70} % SpaceInvaders
& & --- & \multicolumn{1}{c}{0.67} % Freeway
\\ \bottomrule
\end{tabular}
}
\end{sc}
\end{small}
\end{center}
\vskip -0.1in
\end{table*}

\subsection{Experimental Setup}\label{sec:experimental-setup}

To validate the proposed method, we conducted experiments comparing CompoNet with other CRL methods from the literature across three sequences of tasks.
The first sequence includes 20 robotic arm manipulation tasks (a sequence of 10 different tasks repeated twice) from Meta-World \citep{yu2020meta}, an established benchmark in meta-learning and multi-task learning communities. In these tasks, states and actions consist of low dimensional real-valued vectors and a budget of $\Delta = 1M$ timesteps per task has been used.\footnote{
We utilize the same task sequence as in the CW20 benchmark proposed by \citet{wołczyk2021continual}, using the second version of the tasks from \citet{yu2020meta} due to code deprecation issues. See Appendix~\ref{sec:mw-vs-cw} for further details.} 
Following the common practice of the literature, the Soft Actor-Critic (SAC) \citep{haarnoja2018soft} algorithm has been used to optimize every method. 
%%%
The other two task sequences are selected from the Arcade Learning Environment~\citep{machado18arcade}. 
% In this case, states consist of $210\times160$ RGB images and actions are discrete.
\modi{In this case, actions are discrete, and states consist of RGB images of $210\times160$ pixels. Thus, we employ a CNN encoder as described in Section~\ref{sec:encoder} to encode images into a lower dimensional space (see Appendix~\ref{sec:appendix-encoder}).}
The first sequence corresponds to the 10 playing modes of the \textit{ALE/SpaceInvaders-v5} environment, while the last one to the 7 playing modes of the \textit{ALE/Freeway-v5} environment, with $\Delta = 1M$ timesteps per task. The Proximal Policy Optimization (PPO) \citep{schulman2017proximal} algorithm is employed to optimize the methods in these tasks as commonly employed in the literature~\citep{huang2022cleanrl}.

\paragraph{Methods.} 
%\subsubsection{Methods}
%\textbf{Methods.} 
In addition to CompoNet, we include five methods for comparison. 
% Baseline
The baseline involves training a randomly initialized NN for each task, serving as the baseline reference. We expect CRL methods to perform at least as well as the baseline.
% , otherwise indicating that the method suffers from interference. 
Adhering to common practice in the literature \citep{wołczyk2021continual,wolczyk2022disentangling}, FT-1 continuously fine-tunes a single NN model across all tasks; expecting more advanced CRL methods to at least match its forward transfer. 
FT-N adopts a similar approach but preserves the models trained at each task to prevent forgetting.
%% As commonly considered in the literature \citep{wołczyk2021continual,wolczyk2022disentangling}, FT-1 entails continuously fine-tuning a single NN model from the first to the last task, expecting more complex CRL methods to match its forward transfer. FT-N employs the same approach but stores trained models at each task to avoid forgetting.
% ProgNet
ProgressiveNet~\citep{rusu2016progressive} is selected for its similarity to this work and for serving as a basis for more complex approaches. This method instantiates a new NN module every time the task changes, freezing the parameters of previous modules and adding lateral connections between the hidden layers of the modules. Finally, 
%%%% He tenido que meter el mbox para que no partiera la cita en dos y haer clicable la tabla
\mbox{PackNet \citep{mallya2018packnet}}, stores the parameters to solve every task of the sequence in the same network by building masks to avoid overwriting the ones used to solve previous tasks. PackNet has shown strong results in previous works~\citep{nips2019compacting,wolczyk2022disentangling}.
The complete description of each environment and task is available in Appendix~\ref{sec:envs-and-tasks}, while implementation details and hyperparameters are provided in Appendix~\ref{sec:implementation-details}.

\subsection{Results} \label{sec:perf-and-ft}

Table~\ref{table:main-results} shows the results for all methods in each of the task sequences under the metrics described in Section~\ref{sec:metrics}, including the RT for every sequence (computed from the transfer matrices in Appendix~\ref{sec:fwd-transf-mats}).

% Performance
% In terms of performance, and in the case of the Meta-World sequence, both growing NN methods are the ones with the highest performance, although CompoNet still outperforms ProgressiveNet. In the SpaceInvaders sequence, CompoNet and FT-N are the methods with the highest performance, achieving the same score. ProgressiveNet is the method with the third best performance obtaining a 28\% lower score the the latter ones. In the final sequence, Freeway, CompoNet significantly outperforms the rest of the methods by a significant margin, followed by FT-N with 13\% lower performance.
Regarding performance in the Meta-World sequence, both growing NN methods exhibit the highest performance, with CompoNet outperforming ProgressiveNet. For the SpaceInvaders sequence, CompoNet and FT-N show the highest performance, achieving identical scores, while ProgressiveNet trails with a 0.28 lower score. In the Freeway sequence, CompoNet significantly outperforms all other methods by a considerable margin, followed by FT-N with 0.13 lower performance.

% achieving an score of 0.94, while the second highest score is 0.81 by FT-N.  
% Regarding the rest of the , we observe that ProgressiveNet is the second method with the best overall results, while PackNet and fine-tuning suffer from significantly lower performance, even having lower average performance than training from scratch in some task sequences.   

% Forward transfer
In terms of forward transfer, CompoNet stands out as the only method achieving positive forward transfer (i.e. has no interference) in the challenging Meta-World sequence. This highlights the robustness of CompoNet given the high interference nature of the sequence demonstrated by the negative RT value. In SpaceInvaders, CompoNet achieves the highest forward transfer, followed by FT-1 and FT-N, the only three methods surpassing the RT in this sequence. Finally, in the Freeway sequence, CompoNet notably outperforms other methods and the RT. Across all sequences, CompoNet consistently improves the RT, 
demonstrating that its knowledge transfer capabilities go beyond fine-tuning the previous task of highest transfer, also leveraging knowledge composition.

% \modi{Note that although the forward transfer of the baseline method is zero}

% In terms of forward transfer, CompoNet is the only method that obtains positive forward transfer (i.e. has no interference) in Meta-World, a challenging sequence for its high interference shown by the negative RT value, evidencing the robustness of the presented method. 
% Observing SpaceInvaders, CompoNet is the method with the highest forward transfer, followed by FT-1 and FT-N. Note that the three methods exceed the RT. 
% In the Freeway sequence, CompoNet reaches specially higher forward transfer values compared to the rest of the methods and the RT. 
% Note that in all three sequences, CompoNet improves the RT, demonstrating that the knowledge transfer properties of the method go beyond fine-tuning on the previous task of highest transfer, and that leverages knowledge composition.

% % Forgetting
% With respect to forgetting, CompoNet, FT-N, ProgressiveNet, and PackNet naturally avoid the issue by storing the parameters learned in each task. In the case of the rest of the methods, forgetting is reported in Table~\ref{table:all-forgs} in the appendix.

Refer to Appendix~\ref{sec:success-rate-curves} for the success rate curves in each task. 
In Appendix~\ref{sec:forgetting-appendix} we describe and provide results for forgetting, another common metric in the CRL literature. Note that this metric has been omitted from this section as the only affected methods are the baseline and FT-1.
% Results on forgetting are omitted from this section and are provided in Appendix~\ref{sec:forgetting-appendix}, as the only methods affected by this issue are the baseline and FT-1.

% With respect to the forgetting metric also employed in the CRL literature, we define a provide results for this metric in Appendix~\ref{sec:further-results}, as CompoNet naturally avoids forgetting under the employed assumptions.

% Regarding the forgetting metric used in CRL literature, we define and present results for this metric in Appendix~\ref{sec:further-results}, as CompoNet naturally mitigates forgetting under the employed assumptions.

% Summary
%In summary,  CompoNet achieves the highest performance and and forward transfer regardless of the task sequence, while not suffering from forgetting.

\begin{figure*}[ht!]
     \centering
     \begin{subfigure}{0.47\textwidth}
         \centering
         \includegraphics[width=\textwidth]{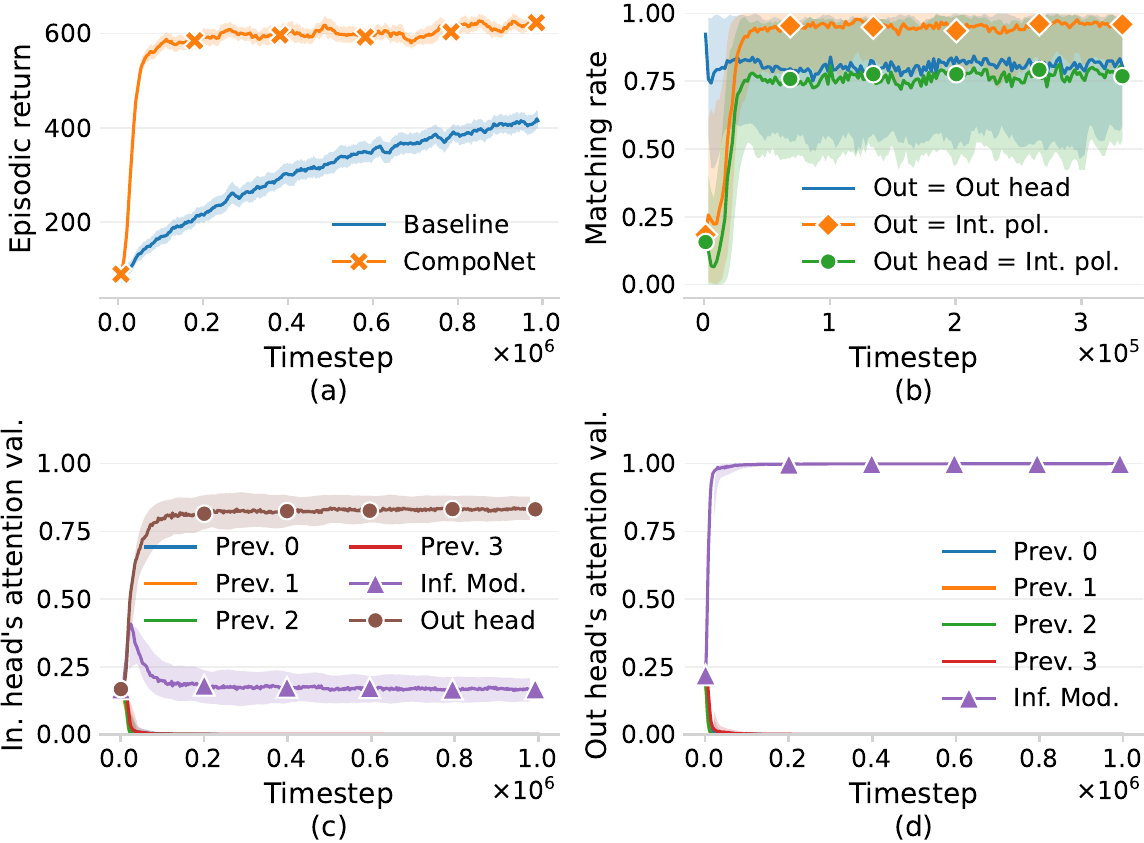}
         % \caption{}
         %\label{fig:arch-val-obj-1}
     \end{subfigure} 
          \hfill
          {\color{lightgray} \vrule}
          \hfill
     \begin{subfigure}{0.47\textwidth}
         \centering
         \includegraphics[width=\textwidth]{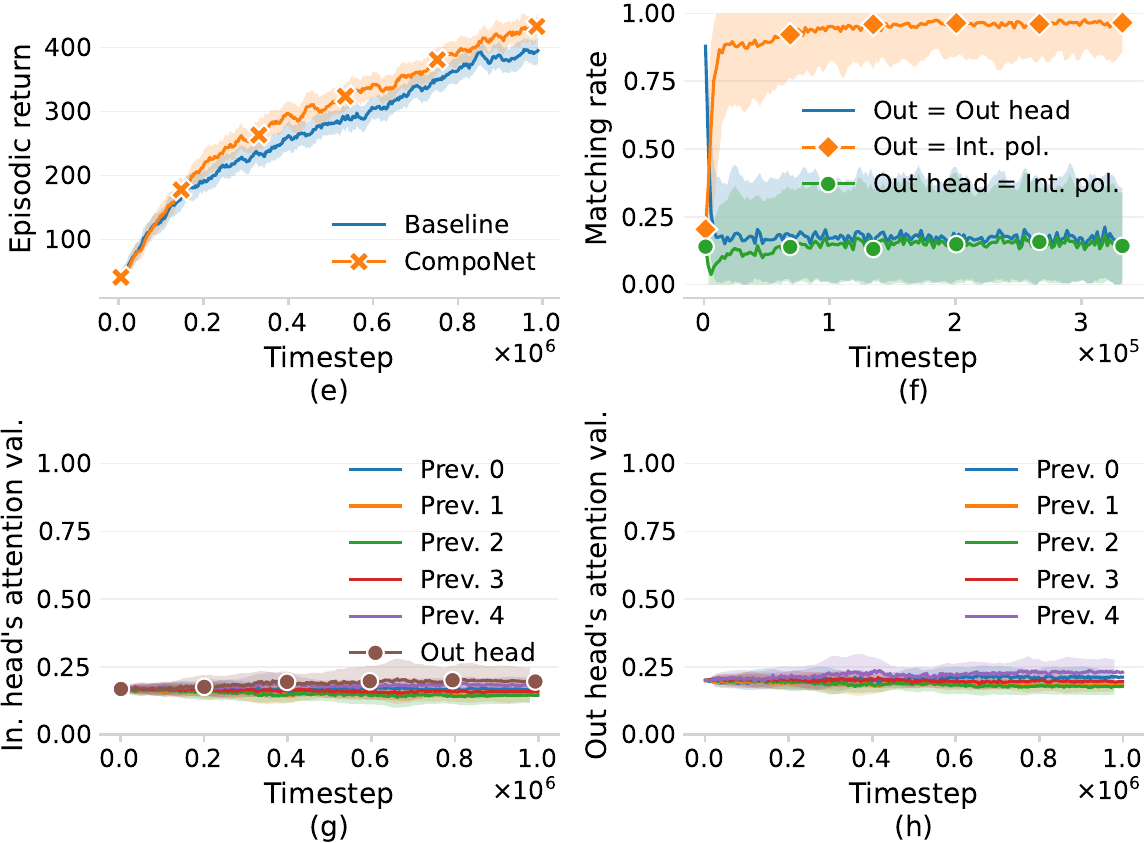}
         % \caption{}
         %\label{fig:arch-val-obj-3}
     \end{subfigure}
     \caption{
     Empirical results on the fulfillment of objectives (i) and (iii) from Section~\ref{sec:model} that motivated the design of CompoNet. In the leftmost figures, CompoNet is trained on the fifth task of SpaceInvaders with four non-informative previous policies that sample their output from a uniform Dirichlet distribution, and one policy trained to solve the current task (\textit{Inf. Mod.}). In the rightmost figures, CompoNet is trained on the sixth task of SpaceInvaders with five non-informative previous policies. Results aggregate 10 random seeds. The matching rate indicates the frequency with which the action of the highest probability from the output of the last module, or some of its components, matches with each other. For clarity, the X axis of Figures~\ref{fig:arch-val}.b and~\ref{fig:arch-val}.f are shortened to the first $3\times10^5$ timesteps.  
     % Empirical results on the fulfillment of objectives ii and iii from Section~\ref{sec:model} that motivated the design of CompoNet. In (a), CompoNet is trained in the fifth task of SpaceInvaders with four non-informative previous modules that sample their output from a uniform Dirichlet distribution, and one that has been trained to solve the current task (\textit{Inf. Mod.}). In (b), CompoNet is trained in the sixth task of SpaceInvaders with five non-informative previous modules. Results aggregate the information of 10 random seeds. Matching rate refers to the rate in which the action of highest probability of the output of the last module or some its components matches with each other. For visibility purposes, the X axis of Figure~\ref{fig:arch-val-obj-1}.ii and~\ref{fig:arch-val-obj-3}.ii have been shortened.  
     }
     \label{fig:arch-val}
\end{figure*}

\subsection{Architectural Validation}\label{sec:archi-val}

Returning to the scenarios presented in the first lines of Section~\ref{sec:model}, in (i) CompoNet should be able to efficiently reuse previous policies; in (ii) it should extract useful information from previous policies to accelerate the learning process on the current task; \modi{and in (iii),} when previous tasks have no relation with the current one, it should learn a policy from scratch. 
The ability of CompoNet to leverage forward knowledge transfer to more efficiently solve unseen tasks (second scenario) has been demonstrated in Section~\ref{sec:perf-and-ft}, \modi{here,} we verify that CompoNet fulfills the expected behavior in the first and last scenarios.

% Objective 1
Regarding the first objective, we train CompoNet on the fifth task of the SpaceInvaders sequence with a previous policy already trained on this task (referred to as \textit{Inf. Mod.} \modi{in Figure~\ref{fig:arch-val}}) and four random modules that output a random policy sampled from a uniform Dirichlet distribution at every timestep. 
% episodic return
Figure~\ref{fig:arch-val}.a shows the episodic return\footnote{Episodic return is defined as the sum of all rewards for an episode (e.g., time until the game is over).} curves (greater is better), where CompoNet clearly benefits from the information of the previous policy trained in the current task.
% Output attention head
Observing the behavior of the output attention head in Figure~\ref{fig:arch-val}.d, we see it assigns high attention scores to the previous policy trained in the task at hand (\textit{Inf. Mod.}, marked with triangles), letting the output from this policy to pass through the attention head to the other blocks of the current module.
% Input attention head
In turn, the input attention head (see Figure~\ref{fig:arch-val}.c) attends to the previous policy that solves the task and to the output attention head, whose output is equal to the output of the previous policy that solves the current task at this point on the training.
% Matches
Finally, Figure~\ref{fig:arch-val}.b shows the rate at which the action of the highest probability of two outputs are equal, computed for the output of the internal policy (\textit{Int. Pol.}), output attention head (\textit{Out head}), and the final output of the model (\textit{Out}).
In the initial timesteps, the output of the model matches with the result of the output attention head, and the internal policy is barely used.
After several timesteps, the internal policy learns to imitate the result of the output attention head, which is mostly used as the final output of the model.

% Objective 3
To validate CompoNet in the last scenario, in the rightmost plots of Figure~\ref{fig:arch-val} CompoNet is trained to solve the sixth task of SpaceInvaders, having five previous modules that output a random policy in the same way as in the previous experiment. 
Observing Figure~\ref{fig:arch-val}.e, we see that CompoNet matches (and slightly improves, possibly due to the extra stochasticity introduced by the previous policies) the performance of training a policy from scratch.  
Moreover, Figure~\ref{fig:arch-val}.f shows that the final output of the model is completely determined by the internal policy after a few training steps, effectively overwriting the result of the output attention head.
Finally, in Figures~\ref{fig:arch-val}.g and~\ref{fig:arch-val}.h we can see that both attention heads give uniform attention to all of their inputs. This behavior is expected as none of the previous modules provides any useful information for the task at hand.

As additional evidence on the validity of the architecture described in Section~\ref{sec:self-compo-pol}, Appendix~\ref{sec:ablations} presents ablation studies to verify the functionalities of the different blocks that constitute the architecture of the self-composing policy module. \modi{Finally, Appendix~\ref{sec:att-scalability} demonstrates the scalability of CompoNet to handle hundreds of previous modules.} 

\section{Conclusions}

% Good performance
We propose CompoNet, a modular growing NN architecture for CRL that naturally avoids catastrophic forgetting and interference while leveraging the knowledge obtained in previous tasks to address new ones, as demonstrated in experiments across task sequences of different natures.
% Scalability
Unlike other growable NN approaches, CompoNet grows linearly in the number of parameters with the number of tasks and shows substantially better scalability in terms of inference.
These properties allow CompoNet to scale to many sequential tasks in the practice, without sacrificing plasticity and performance.
Contrary to most compositional NN methods, CompoNet does not require a dedicated model to compose learned modules, as modules learn to compose themselves.
Moreover, experiments demonstrate that CompoNet is able to utilize the knowledge obtained in previous tasks while learning without interference when the current task is unrelated to prior ones.
% Moreover, experiments show that CompoNet extracts information from modules learned in previous tasks when possible while being able to learn without interference when the current task has no relation with the previous ones.

% Limitations
We believe that this work is a step forward in the development of CRL agents that benefit from learning from numerous sequential tasks in the practice, however, several challenges remain.
% We believe that this work is a step forward in the development of agents that could learn from numerous of sequential tasks in the practice, leveraging forward transfer while mitigating forgetting and interference. 
% However, several challenges remain.
% Complexity
% The memory complexity of CompoNet is linear, and the cost of inference is quadratic. 
Although CompoNet already scales to many tasks in practice, \modi{improving its computational cost is key for 
considering never-ending continual learning problems.}
% future work. 
We think that quantization~\citep{xiao2023smoothquant} and policy distillation~\citep{rusu2015policy} could allow this issue to be addressed. 
% Finally, as with many other works in the CRL literature, this work assumes that the boundaries between tasks and their identifiers are known to the agents. We believe that future work on methods that autonomously extract such information is an important milestone for the future of CRL research.
Finally, likewise to most works in the CRL literature~\citep{rusu2016progressive,nips2019compacting,wołczyk2021continual,wolczyk2022disentangling}, our work assumes agent knowledge of task boundaries and identifiers.
\modi{Incorporating a method to extract such information in CompoNet would allow 
new modules to be automatically added.
% automatically adding of new modules when needed. 
Although this is beyond the scope of this work, we believe that future research on systems that extract such information is pivotal for advancing CRL research.}
% We believe that future research focusing on methods that autonomously extract such information is pivotal for advancing CRL research. 

%\textcolor{red}{\textbf{[Reduce hype]} To the best of our knowledge, this work has been the first one to employ a pretrained visual foundation model for the sake of training RL agents, significantly reducing the memory cost per task of CompoNet, and enabling knowledge transfer across visually different games.}

% \paragraph{Limitations} \textcolor{red}{[\textbf{TODO} Task transition knowledge required, cuadratic compu cost]} Although CompoNet significantly improves the memory cost of existing growable NNs, this could be considerably reduced by detecting and removing modules that are not being attended by any other module. 
% The latter would be analogous to how humans forget skills that we do not regularly use. 
% Finally, we think that a logical next step for our work is to detect transitions between tasks (e.g. via world models) and automatically add new modules to the architecture.

\section*{Acknowledgements}

We are grateful to Jose A. Pascual for the technical support. We also thank Jon Vadillo and Ainhize Barrainkua for reading the preliminary versions of the paper.

Mikel Malagón acknowledges a predoctoral grant from the Spanish MICIU/AEI with code PREP2022-000309, associated with the research project PID2022-137442NB-I00 funded by the Spanish MICIU/AEI/10.13039/501100011033 and FEDER, EU.
 
This work is also funded through the BCAM Severo Ochoa accreditation CEX2021-001142-S/MICIN/AEI/10.13039/501100011033; and the Research Groups 2022-2025 (IT1504-22), and Elkartek (KK-2021/00065 and KK-2022/00106) from the Basque Government.

\section*{Impact Statement}

This paper presents work whose goal is to advance the field of Machine Learning. There are many potential societal consequences of our work, none which we feel must be specifically highlighted here.

\bibliography{references}
\bibliographystyle{icml2024}

%%%%%%%%%%%%%%%%%%%%%%%%%%%%%%%%%%%%%%%%%%%%%%%%%%%%%%%%%%%%%%%%%%%%%%%%%%%%%%%
%%%%%%%%%%%%%%%%%%%%%%%%%%%%%%%%%%%%%%%%%%%%%%%%%%%%%%%%%%%%%%%%%%%%%%%%%%%%%%%
% APPENDIX
%%%%%%%%%%%%%%%%%%%%%%%%%%%%%%%%%%%%%%%%%%%%%%%%%%%%%%%%%%%%%%%%%%%%%%%%%%%%%%%
%%%%%%%%%%%%%%%%%%%%%%%%%%%%%%%%%%%%%%%%%%%%%%%%%%%%%%%%%%%%%%%%%%%%%%%%%%%%%%%
\newpage
\appendix
\onecolumn

\counterwithin{figure}{section}
\counterwithin{table}{section}

\section{Vision Foundation Models for Visual Control Tasks in RL}\label{sec:dino-appendix}

In the context of Section~\ref{sec:encoder}, where different state encoding strategies for CompoNet are discussed, in this appendix we show preliminary results on the possibility of using a single pre-trained visual foundation model for generating low dimensional representations of high dimensional images. 
% In order to increase the scalability of CompoNet in visual tasks, a pre-trained vision model can be used to extract useful visual features for all the modules of CompoNet. In consequence, e

% By employing a single encoder, the number of parameters required by each module of the CompoNet architecture could be considerably reduced in some cases. 
By employing a single encoder, the number of parameters required by each module of the CompoNet architecture could be considerably reduced in some cases.  In fact, this strategy can be of special interest in task sequences where states are composed of high-dimensional images or in extremely large sequences, where the cost of maintaining a CNN for each module is unaffordable. 
Note that this is not the case for the image-based sequences considered in this work (SpaceInvaders and Freeway) and that this appendix only aims to shed some light on alternative strategies to follow in the mentioned hypothetical cases.

% In this appendix we show preliminary results on the possibility to use a single, pre-trained, visual foundation model for generating low dimensional representations of high dimensional images. 
Specifically, for this experiment, we employ the recent DINOv2 vision foundation model by \citet{oquab2023dinov2}, which demonstrated the ability to generate all-purpose visual features from images that can be used for multiple tasks such as image classification, segmentation, or depth recognition. 

For this appendix, we have employed the smallest model from \citet{oquab2023dinov2}, a 21M parameter vision transformer \cite{dosovitskiy2020image} with an embedding size of 384 (referred to as $d_\text{enc}$ in Sections~\ref{sec:encoder} and onwards) and a patch size of 14 pixels.
As the model is a vision transformer, its input consists of a sequence of non-overlapping patches of the input image and a \textit{CLS} token. The output is a sequence of the same length that includes patch tokens and the class token (corresponding to \textit{CLS}), that serves as the representation of the whole image. Consequently, for an image of $224\times 224$ pixels, the DINOv2 ViT-S model outputs a single class token and 256 patch tokens, where the dimension of each one is 384.

\begin{figure}[h!]
    \centering
    \begin{subfigure}{0.47\textwidth}
        \includegraphics[width=\textwidth]{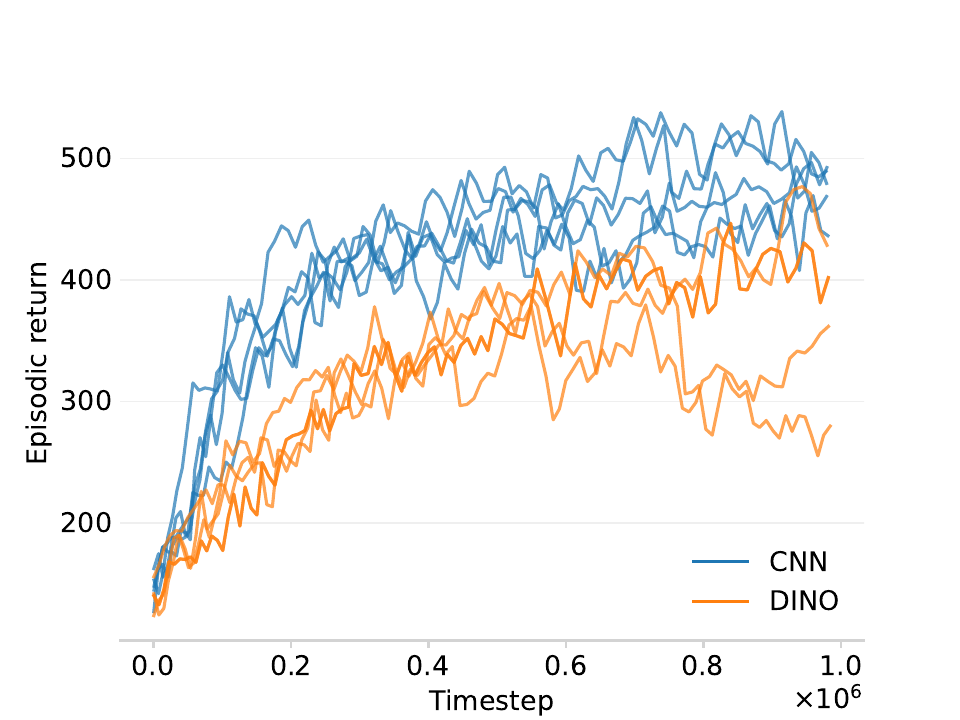}
        \caption{}\label{fig:dino-vs-cnn-return}
    \end{subfigure}
    \begin{subfigure}{0.38\textwidth}
        \centering  
        \includegraphics[width=0.7\textwidth]{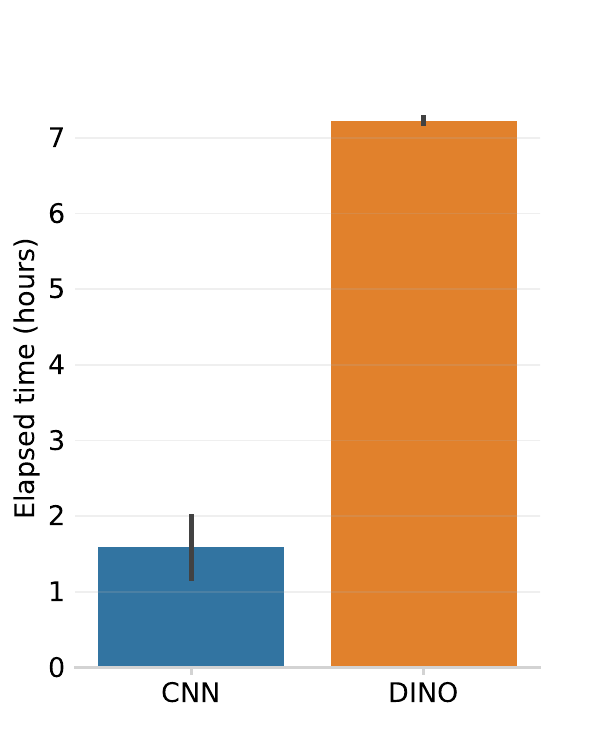}
        \caption{}\label{fig:dino-vs-cnn-time}
    \end{subfigure}
    \caption{Comparison of training an agent in the first task of the SpaceInvaders sequence using the representations generated by a pre-trained DINO model, and a CNN encoder that is trained with the agent. 
    The leftmost figure compares both approaches in terms of episodic return, while the rightmost compares the total execution times to reach 1M timesteps.
    Both figures contain the results from 5 runs per setting.}
    \label{fig:dino-vs-cnn}
\end{figure}

Figure~\ref{fig:dino-vs-cnn} shows the comparison of training an agent (the baseline method using the hyperparameters from Table~\ref{table:hyperparams-ppo}) to solve the first task of the SpaceInvaders sequence using the representations encoded by the DINOv2 pre-trained model, and a CNN that is trained \modi{alongside} the agent. 
% As depicted, the final episodic return obtained by both approaches is similar, although the CNN-based agent can reach higher scores faster. 
\modi{As depicted, in most cases, the final episodic return obtained by both approaches is similar. However, the CNN-based agent can reach higher scores faster and the DINO-based agent obtains considerably lower values for a couple of repetitions.}
Although the analysis is preliminary and results can be improved, the similarity in the final performance demonstrates that the representations generated by the pre-trained DINO model can be used to successfully train an RL agent to solve this type of visual task. Moreover, note that the DINO model has not been trained with the RL agent, and despite not having data from this game in its training, it is still able to extract useful information to solve the task at hand.    

Remember that using \modi{visual foundation models such as DINO,} would only be needed in extremely large task sequences, or in tasks where states are composed of considerably high dimensional images. 
In fact, the computational cost of using such a large encoder model (the smallest DINO model has 21M parameters) is counter-productive for the small image-based tasks considered in this work.
This is supported by the training time comparison presented in Figure~\ref{fig:dino-vs-cnn-time}, where the training times of both approaches are shown. As depicted, the CNN-based models require less than half of the training time of the DINO-based ones.

\section{Memory Cost of CompoNet}\label{sec:memory-cost-appendix}

Memory cost has been a major concern in this work, as it is one of the major issues of growing NNs, greatly limiting their scalability. In this section, we show that the presented architecture grows linearly in the number of parameters with respect to the number of tasks, being scalable in terms of memory.

CompoNet requires adding a new policy module every time a new task arrives, however, as described in Section~\ref{sec:self-compo-pol} and illustrated in Figure~\ref{fig:compo-unit}, each module only consists of a few components that require parameters.
Specifically, six linear transformations and a feed-forward block, the rest of the operations have no parameters.
As specified in Section~\ref{sec:self-compo-pol}, the size of the matrices that correspond to these linear transformations is determined by the output dimension of the encoder $d_\text{enc}$, the hidden vector size of the model $d_\text{model}$ and the number of actions $|\Actions|$. Additionally, the number of parameters of the feed-forward block is only dependent on the number of layers and the hidden vector dimension (assumed to be equal to $d_\text{model}$).
Note that the size of the matrices used in the attention heads and the feed-forward network are only dependent on fixed hyperparameters. Therefore,
the number of parameters of a self-composing policy module is constant \modi{and independent of the number of tasks addressed}. % with respect to the number of tasks.
Considering $m$ to be the number of parameters of a self-composing module, the number of parameters of the CompoNet architecture is $\mathcal{O}(m\cdot n)$, where $n$ is the number of tasks. 
As $m$ is constant, the memory complexity of CompoNet with respect to the number of tasks is linear, $\mathcal{O}(n)$.

\begin{figure}[h]
    \centering
    \includegraphics[width=0.45\textwidth]{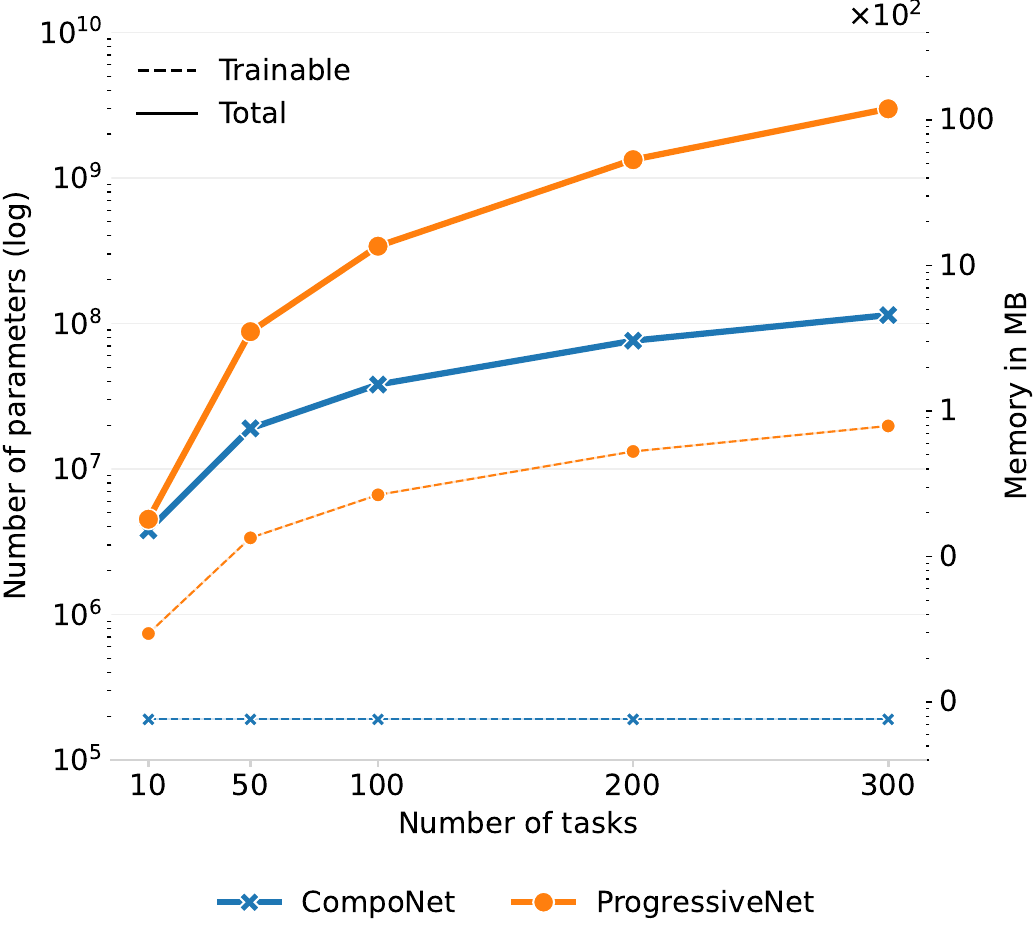}
    \caption{Growth in memory of CompoNet and ProgressiveNet models as the number of tasks (assuming an NN module per task) increases. \modi{The count is given in the total number of parameters, depicted with solid lines, and in the number of trainable parameters (not frozen), in dashed lines.} Note that we assume 32-bit floats are used to represent the parameters of the models. Hyperparameters correspond to the ones utilized in the Meta-World sequence: $d_\text{enc} = 39$, $d_\text{model} = 256$, and $|\mathcal{A}| = 4$.}
    \label{fig:memory-cost}
\end{figure}

Figure~\ref{fig:memory-cost} provides a comparison of the growth of CompoNet and ProgressiveNet in the number of parameters and the \modi{required memory} in Megabytes. The hyperparameters are set according to those employed in the Meta-World sequence (detailed in Appendix~\ref{sec:implementation-details} and Table~\ref{table:hyperparams-sac}).
% \modi{Besides the hidden dimension of the models, the} rest of the hyperparameters are set according to the ones employed in the Meta-World sequence (detailed in Appendix~\ref{sec:implementation-details} and Table~\ref{table:hyperparams-sac}).

\begin{figure}[h]
    \centering
    \includegraphics[width=0.5\textwidth]{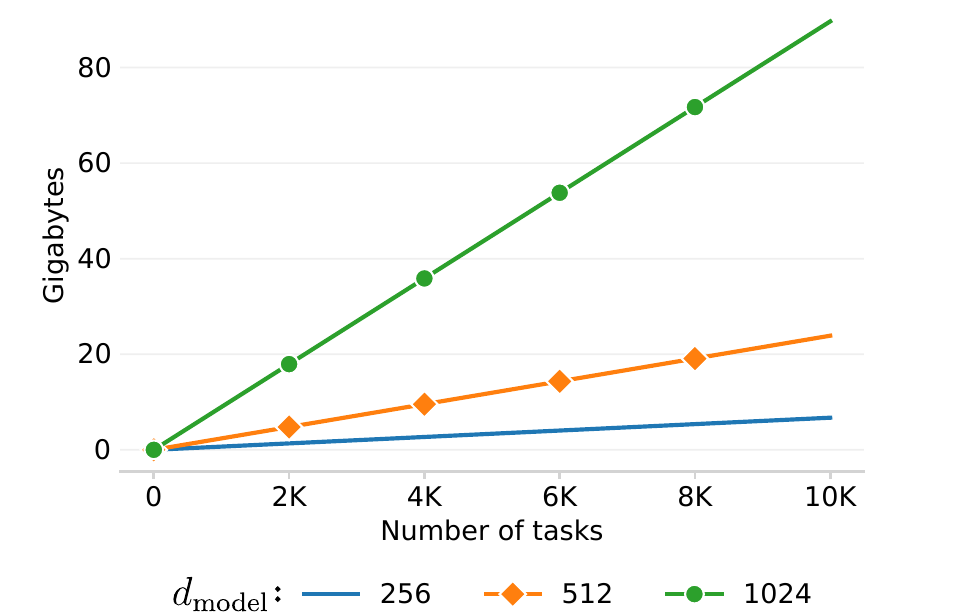}
    \caption{Memory size in Gigabytes required by CompoNet with respect to the number of tasks for different $d_{\text{model}}$ values. Hyperparameters correspond to the ones from the Meta-World sequence: $d_\text{enc} = 39$, and $|\mathcal{A}| = 4$.}
    \label{fig:memory-cost-large}
\end{figure}

To get a sense of the possible scalability to extremely large task sequences, Figure~\ref{fig:memory-cost-large} shows the memory cost (in gigabytes) of CompoNet for up to 10K tasks and different $d_\text{model}$ values. As shown, in the case of $d_\text{enc}=256$ (which corresponds to the one used in the experimentation) the proposed architecture can grow to more than 10k modules in a single NVIDIA A5000 GPU with 24GB of VRAM memory.

\section{Computational Cost of Inference in CompoNet}\label{sec:cost-inference}

% In this section, we discuss the computational cost of running inference in the proposed CompoNet architecture with respect to the number of tasks.
\modi{In this section, we analyze the computational cost of inference with respect to the number of tasks for the CompoNet architecture.
First, we describe the computational complexity of the model in big-$O$-notation. Then, we study the empirical computational cost of inference of CompoNet for hundreds of tasks.}

\subsection{Computational Complexity of Inference}\label{sec:complexity-inference}

According to the diagram presented in Figure~\ref{fig:compo-unit}, the self-composing policy module can be divided into three parts: the output attention head, the input attention head, and the internal policy. In the following lines, we respectively develop the computational complexity of each of the three parts with respect to the number of self-composing policy modules ($n$).\footnote{\modi{The computational complexity of multiplying an $n\times m$ and an $m\times p$ matrix is $O(nmp)$.}}

\paragraph{Output Attention Head.} The input of this block comprises the current state representation $\mathbf{h_s}$, the matrix with the results of the previous modules $\Phi^{k;\mathbf{s}}$, and the same matrix with the positional encoding $\Phi^{k;\mathbf{s}} + E_\text{out}$. 
The first element, $\mathbf{h_s}$, has a constant dimension $d_\text{enc}$, while the $\Phi^{k;\mathbf{s}}$ matrix has a size of $(n-1) \times a$, where the second dimension $a$ is constant with respect to the number of modules $n$ but the first is not.
This implies that the computational cost of calculating the key $K$ matrix and the dot product required to compute the attention
depends on the number of modules $n$.~\footnote{
Remember from Section~\ref{sec:model} that $\bm{q} = \bm{h_s} W_\text{out}^Q$, $K = (\Phi^{k;\bm{s}} + E_\text{out}) W_\text{out}^K$, and $V = \Phi^{k;\bm{s}}$ where $W_\text{out}^Q\in\mathbb{R}^{d_\text{enc}\times d_\text{model}}$, $W_\text{out}^K \in \mathbb{R}^{|\Actions| \times d_\text{model}}$, and $E_\text{out}$ is a positional matrix of the same size as $\Phi^{k;\bm{s}}$. Moreover, attention is computed as $\text{Attention}(\bm{q}, K, V) = \text{softmax}\left( qK^T / \sqrt{d_\text{model}} \right) V$. 
} Formally,
\begin{equation}
    T_\text{input}(n) = \underbrace{d_\text{enc} \cdot d_\text{model}}_\text{Compute $\mathbf{q}$} + \underbrace{(n-1)\cdot a \cdot d_\text{model}}_\text{Compute $K$} + \underbrace{d_\text{model}\cdot (n-1)}_{\mathbf{q}K^T} + \underbrace{O(n-1)}_\text{Cost of softmax} + \underbrace{a\cdot (n-1)}_\text{Mult. att. and $V$}
\end{equation}
% To the contrary of the internal policy,
\modi{Note that all the terms} in $T_\text{input}(n)$ except the first depend on $n$. As there is no term with higher order than $n$, then the computational complexity of the output attention head is $T_\text{input}(n) = O(n)$.

\paragraph{Input Attention Head.} This block requires the representation of the current state $\bm{h_s}$, and the row-wise concatenation of the result from the previous block with $\Phi^{k;\bm{s}}$. Note that similarly to the output attention head, this block also adds a positional encoding to the keys matrix, however, the values matrix ($V$) is computed via a linear transformation (unlike the previous block, where $V=\Phi^{k;\bm{s}}$). In this case, the computational complexity is, 
\begin{equation}
    T_\text{output}(n) = \underbrace{d_\text{enc} \cdot d_\text{model}}_\text{Compute $\mathbf{q}$} + \underbrace{2\cdot n\cdot a \cdot d_\text{model}}_\text{Compute $K$ and $V$} + \underbrace{d_\text{model}\cdot n}_{\mathbf{q}K^T} + \underbrace{O(n)}_\text{Softmax} + \underbrace{n\cdot d_\text{model}}_\text{Mult. att. and $V$}
\end{equation}
Likewise the previous attention head, the highest order in $T_\text{output}(n)$ is $n$, and thus, its computational complexity is $T_\text{output}(n) = O(n)$. 

\paragraph{Internal Policy.} The input of the internal policy consists of a vector of size ($d_\text{enc} + d_\text{model}$) and outputs a vector of length $a = |\mathcal{A}|$. Assuming that the internal policy is constructed from linear layers, its computational complexity is defined as the following,
\begin{equation}
    T_\text{int}(n) = \underbrace{d_\text{model}\cdot(d_\text{enc} + d_\text{model})}_\text{First layer} + \underbrace{(l-2)\cdot d_\text{model}^2}_\text{Intermediate layers} + \underbrace{d_\text{model}\cdot a}_\text{Last layer}
\end{equation}
Note that none of the terms of $T_\text{int}(n)$ depends on $n$ (the number of policy modules). Therefore, the computational cost of the internal policy \modi{is constant and independent of the number of modules (i.e., number of tasks)}, $T_\text{int}(n) = O(1)$.\footnote{For the sake of simplicity, this definition of the internal policy ignores possible activation and normalization layers.} 

\paragraph{Total Complexity.} Considering the computational complexity of inference of computing the three parts that constitute the self-composing policy module, the total complexity of a single module is \modi{$T_\text{output}(n)+T_\text{input}(n)+T_\text{int}(n) = O(n) + O(n) + O(1) = O(n)$.}
However, considering a CompoNet model of $n$ policy modules, obtaining its final output requires computing the results of all $n$ modules sequentially, as each module depends on the previous ones. Consequently, although the complexity of inference for a policy module is linear, the total complexity of CompoNet is $n \cdot O(n) = O(n^2)$.

\subsection{Empirical Computational Cost}

The previous section focused on the theoretical computational complexity of inference for the CompoNet architecture, in this section, we empirically measure the time required to perform these computations in practice. 

Specifically, we run inference in CompoNet and ProgressiveNet models with an increasing number of modules, simulating inference in models that would have faced hundreds of tasks, maintaining a module for each one. 
To provide practical and realistic measurements, we set the same hyperparameters employed in the Meta-World sequence for this experiment (see Appendix~\ref{sec:implementation-details} and Table~\ref{table:hyperparams-sac}). 

\begin{figure}[H]
    \centering
    \includegraphics[width=0.45\textwidth]{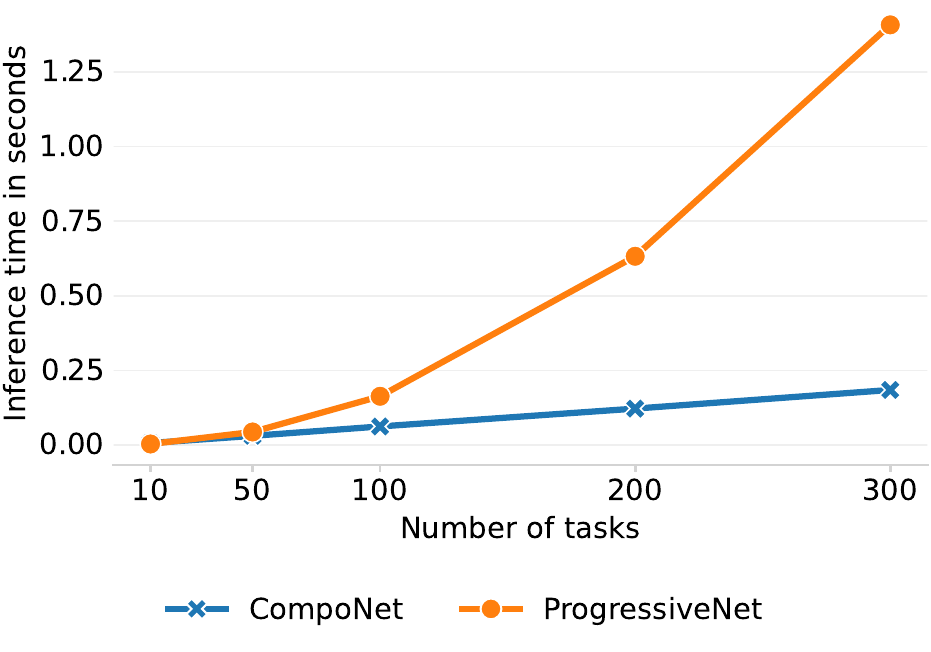}
    \caption{Empirical computational cost of inference of CompoNet and ProgressiveNet with respect to the number of tasks (assuming one module is instantiated for every task). \modi{Results depict the average inference time over one minute of recording. This ensures a minumum of 40 inferences to compute the estimate ($40 = 60 / 1.5$).} 
    %Results depict the average of a minute of total recording per point in the X-axis, ensuring a minimum of 40 inference tests for each of the points in the plot. 
    Hyperparameters are: $d_\text{enc} = 39$, 
    $d_\text{model} = 256$, and $|\mathcal{A}| = 4$. Results were measured in a machine with an AMD EPIC 7252 CPU and an NVIDIA A5000 GPU.}
    \label{fig:inference-cost}
\end{figure}

Results are shown in Figure~\ref{fig:inference-cost}. Although the computational complexity of both approaches is quadratic, the quadratic trend of CompoNet grows substantially slower compared to ProgressiveNet. This might be caused by the fact that ProgressiveNet requires the information of all previous modules in every layer of each module (or NN column as referred to in \citet{rusu2016progressive}). In contrast, although CompoNet also requires the information of all previous modules, the keys and values for the attention heads are computed in parallel.

\section{Environments and Tasks} \label{sec:envs-and-tasks}

\begin{figure}[H]
     \centering
     \begin{subfigure}[b]{0.3\textwidth}
         \centering
         \includegraphics[width=50mm]{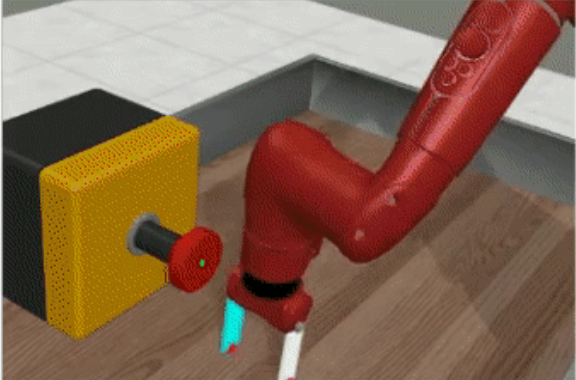}
         \caption{Meta-World}
         \label{fig:env-meta-world}
     \end{subfigure} 
     \begin{subfigure}[b]{0.3\textwidth}
         \centering
         \includegraphics[width=25mm]{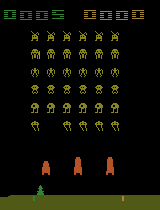}
         \caption{SpaceInvaders}
         \label{fig:env-spaceinvaders}
     \end{subfigure}
     \begin{subfigure}[b]{0.3\textwidth}
         \centering
         \includegraphics[width=25mm]{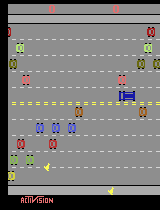}
         \caption{Freeway}
         \label{fig:env-freeway}
     \end{subfigure}
     \caption{\modi{An example of a} frame for each of the considered task sequences.} \label{fig:envs}
\end{figure}

The experimentation is conducted across three task sequences: one consisting of continuous control robotic tasks from Meta-World presented by \citet{yu2020meta}, while the other two correspond to the different playing modes of the SpaceInvaders and Freeway games from the Arcade Learning Environment~\citep{bellemare13arcade,machado18arcade}. An illustration of a task from each sequence is depicted in Figure~\ref{fig:envs}. 

Appendix~\ref{sec:appendix-meta-world}, \ref{sec:appendix-space-invaders}, and \ref{sec:appendix-freeway} describe the environments and tasks of the mentioned sequences. Next, Appendix~\ref{sec:success-scores} provides details on the calculation of the success rate in the SpaceInvaders and Freeway sequences. Finally, Appendix~\ref{sec:fwd-transf-mats} gives the forward transfer matrices used to compute the RT values shown in Table~\ref{table:main-results}.

\subsection{Meta-World}\label{sec:appendix-meta-world}

This sequence is composed of tasks from Meta-world, an open-source suite that includes a variety of robotic arm tasks proposed in~\citet{yu2020meta}. 
Although \citet{yu2020meta} proposed 50 different tasks, in this work we have selected a sequence of 10 tasks, that we repeated twice making 20 tasks in total ($10 + 10$), following the same practice as \citet{wołczyk2021continual}.

In these tasks, \modi{states are 39-dimensional real-valued vectors, $\mathcal{S} \subset \mathbb{R}^{39}$}, where information on the positions, rotations, and states of the different parts of the robotic arm and the objects in the environment are encoded. In turn, the action space consists of a 4-dimensional real-valued vector in the range $[-1, 1]$, which represents the change in the position of the arm and the torque that the gripper fingers should apply.   
The following lines describe the 10 different tasks in order:

\noindent \textbf{hammer-v2.} Hammer a screw into the wall, with random hammer and screw positions every episode.

\noindent \textbf{push-wall-v2.} Bypass a wall and push a puck to a goal, with random positions of the puck and goal in every episode.

\noindent \textbf{faucet-close-v2.} Rotate the faucet clockwise, random faucet position every episode.

\noindent \textbf{push-back-v2.} Push a mug under a coffee machine, with random positions of the mug and the machine every episode.

\noindent \textbf{stick-pull-v2.} Grab a stick and pull a box with the stick, random stick position every episode.

\noindent \textbf{handle-press-side-v2.} Press a handle down sideways, random handle position every episode.

\noindent \textbf{push-v2.} Push the puck to a goal, with random puck and goal positions every episode.

\noindent \textbf{shelf-place-v2.} Pick and place a puck onto a shelf, with random positions of the puck and the shelf every episode.

\noindent \textbf{window-close-v2.} Push and close a window, with a random position of the window every episode.

\noindent \textbf{peg-unplug-side-v2.} Unplug a peg sideways, random peg position every episode.

\subsubsection{A Note on the Selection of the Tasks in the Meta-World Sequence}\label{sec:mw-vs-cw}

The selection of tasks and their order in our Meta-World sequence is directly influenced by the Continual World (CW) benchmark proposed in~\citet{wołczyk2021continual}, specifically the CW20 benchmark, which comprises a sequence of 20 Meta-World tasks (a sequence of 10 different tasks repeated twice) for evaluating CRL methods. 
While we adopt the exact sequence from the benchmark, we use the second version of the environments \modi{available in the up-to-date version of Meta-World at the time of writing} (e.g., \verb|hammer-v1| replaced by \verb|hammer-v2|). This decision is motivated by several factors. 
Firstly, the second version of the library addresses issues present in the first version (e.g., non-Markovian observations in \verb|v1|\footnote{See the following issue for more details: \url{https://github.com/Farama-Foundation/Metaworld/issues/392}.}).
Moreover, the source code of the CW benchmark has not been updated since 2021 and seems to lack maintenance.\footnote{The source code can be found at \url{https://github.com/awarelab/continual_world}. Although the last commit at the time of writing is from 2023, the commit did not modify any code-related files.} Additionally, CW relies on a deprecated version of the \verb|gym| library, with \verb|gymnasium|\footnote{The homepage of the project can be found at \url{https://gymnasium.farama.org/}.} now recommended by \verb|gym| authors as a modern replacement.\footnote{See the README file of the original repository at \url{https://github.com/openai/gym}.} In fact, \verb|gymnasium| is now under the Farama Foundation\footnote{See the homepage of the organization at \url{https://farama.org/}.}, a non-profit organization that promotes the standardization and maintenance of RL libraries. Notably, from the second version of their library, the Meta-World project is also under the Farama Foundation, ensuring ongoing maintenance and sustainability.

\subsection{SpaceInvaders}\label{sec:appendix-space-invaders}

This sequence is based on the \textit{ALE/SpaceInvaders-v5} environment (\modi{see Figure~\ref{fig:env-spaceinvaders}}) of the \verb|gymnasium| library.\footnote{More information at \url{https://gymnasium.farama.org/environments/atari/space_invaders/}.}
In SpaceInvaders, all tasks share the same objective: shoot space invaders before they reach the Earth. The game ends if the player is hit by enemy fire (falling laser bombs) or when invaders reach the ground. Observations consist of $210\times 160$ RGB images of the frames, and actions are: \textit{do nothing}, \textit{fire}, \textit{move right}, \textit{move left}, \textit{combination of move right and fire}, and \textit{combination of move left and fire}. Tasks correspond to the different playing modes of the game. The player has three lives, and destroying space invaders is rewarded (hitting the invaders in the back rows gives more reward). The considered tasks are the following:

\noindent\textbf{Mode 0.} The default playing mode described at the beginning of this section.

\noindent\textbf{Mode 1.} Shields (orange blocks between the player and the invaders, see Figure~\ref{fig:env-spaceinvaders}) move back and forth on the screen, instead of staying in a fixed position. Using them as protection becomes unreliable.

\noindent\textbf{Mode 2.} The laser bombs dropped by the invaders \textit{zigzag} as they come down the screen, making it more difficult to predict the place they are going to land.

\noindent\textbf{Mode 3.} This task combines modes 1 and 2.

\noindent\textbf{Mode 4.} Same as mode 0 but laser bombs fall considerably faster.

\noindent\textbf{Mode 5.} Same as mode 1 but laser bombs fall considerably faster. 

\noindent\textbf{Mode 6.} Same as mode 2 but laser bombs fall considerably faster. 

\noindent\textbf{Mode 7.} Same as mode 3 but laser bombs fall considerably faster. 

\noindent\textbf{Mode 8.} Same as mode 0 but invaders become invisible for a few frames.

\noindent\textbf{Mode 9.} Same as mode 1 but invaders become invisible for a few frames.

\subsection{Freeway}\label{sec:appendix-freeway}

Like the previous sequence, Freeway is based on the \textit{ALE/Freeway-v5} environment from the same library. On Freeway, the objective is to guide a chicken to cross a road with busy traffic (see Figure~\ref{fig:env-freeway}). \footnote{\modi{More information at} \url{https://gymnasium.farama.org/environments/atari/freeway/}.} In the same way as SpaceInvaders, observations consist of the RGB frames of the game. There are only three possible actions: \textit{do nothing}, \textit{move up}, and \textit{move down}. A reward is only given when the chicken reaches the top of the screen after crossing all the lanes of traffic, resulting in an environment of particularly sparse reward. The considered tasks corresponding to the different game modes are the following:

\noindent \textbf{Mode 0.} Corresponds to the default playing mode of the game.

\noindent \textbf{Mode 1.} Traffic is heavier and the speed of the vehicles increases, the upper lane closest to the center has trucks. Trucks are longer vehicles, and thus, more difficult to avoid.

\noindent \textbf{Mode 2.} Trucks move faster than the fastest vehicles of the previous modes, and traffic is heavier.  

\noindent \textbf{Mode 3.} There are trucks in all lanes, and trucks move as fast as in mode 2 in some of the lanes.

\noindent \textbf{Mode 4.} Similar traffic to previous modes, there are no trucks, but the velocity of the vehicles is randomly increased or decreased.

\noindent \textbf{Mode 5.} Same as mode 1 with the speed of the vehicles randomly changing and some vehicles come in groups of two or three very close to each other.

\noindent \textbf{Mode 6.} Same as the previous mode but with heavier traffic.

\subsection{Success Scores for the SpaceInvaders and Freeway Sequences}\label{sec:success-scores}

As described in Section~\ref{sec:metrics}, the CRL metrics we use in this work are defined over the concept of \textit{performance}, which is in turn  defined in terms of the \textit{success rate}. The success rate \modi{is based on the binary metric} that indicates whether the task is solved\textsuperscript{\ref{note:solve}} (1), or not (0). In the case of the Meta-World sequence, the success rate is given by the environments themselves \citep{yu2020meta}. 
As the SpaceInvaders and Freeway sequences are \modi{generated} in this work, we define success as 1 if the episodic return is greater or equal to the \textit{success score}, otherwise zero. 
\modi{In both environments, we aim to have a high success rate for approaches with high performance, and low success rate for lower performing methods. For this purpose,} we compute the success score for each task as 90\% of the average final episodic return \modi{(i.e., the performance once trained)} of all 8 methods (10 random seeds per method).
For reproducibility, the success scores are given in Tables~\ref{table:success-scores-space-inv} and~\ref{table:success-scores-freeway}. Note that in the case of both sequences, the success score is considerably greater than the episodic return obtained by an agent that acts randomly\footnote{Whose actions are always sampled from a uniform distribution over $\Actions$.}, 148.0 and 0.0 for SpaceInvaders and Freeway respectively~\citep{badia2020agent57}. This ensures that a successful policy must at least learn a more sophisticated behavior than a completely random policy.

\begin{table}[H]
\centering
\caption{Success score values for the SpaceInvaders and Freeway task sequences. If the episodic return in a task is greater or equal to this score, the task is considered solved and success is 1, otherwise 0.}
\label{table:success-scores}
%%%%%%%%%%%%%%%%%%%%%%%%%%%%%%%%%%%%%%%
\begin{subtable}{\textwidth}
\centering
\caption{SpaceInvaders}\label{table:success-scores-space-inv}
\begin{small}
\begin{sc}
\begin{tabular}{lcccccccccc}\toprule
 & Task 0 & Task 1 & Task 2 & Task 3 & Task 4 & Task 5 & Task 6 & Task 7 & Task 8 & Task 9 \\ \cmidrule{2-11}
Success score & 340.94 & 366.762 & 391.16 & 386.99 & 379.41 & 383.73 & 393.83 & 367.98 & 484.23 & 456.19 \\ \bottomrule
\end{tabular}
\end{sc}
\end{small}
\end{subtable} \\ \vspace{5mm}
%%%%%%%%%%%%%%%%%%%%%%%%%%%%%%%%%%%%%%%
\begin{subtable}{\textwidth}
\centering
\caption{Freeway}\label{table:success-scores-freeway}
\begin{small}
\begin{sc}
\begin{tabular}{lcccccccc}\toprule
 & Task 0 & Task 1 & Task 2 & Task 3 & Task 4 & Task 5 & Task 6 & Task 7 \\ \cmidrule{2-9}
Success score & 16.65 & 15.1 & 8.27 & 17.09 & 18.54 & 9.43 & 9.14 & 13.96 \\ \bottomrule
\end{tabular}
\end{sc}
\end{small}
\end{subtable} \\ \vspace{5mm}
\end{table}

\subsection{Forward Transfer Matrices}\label{sec:fwd-transf-mats}

\begin{figure}[H]
     \centering
     \begin{subfigure}[b]{0.47\textwidth}
         \centering
         \includegraphics[width=\textwidth]{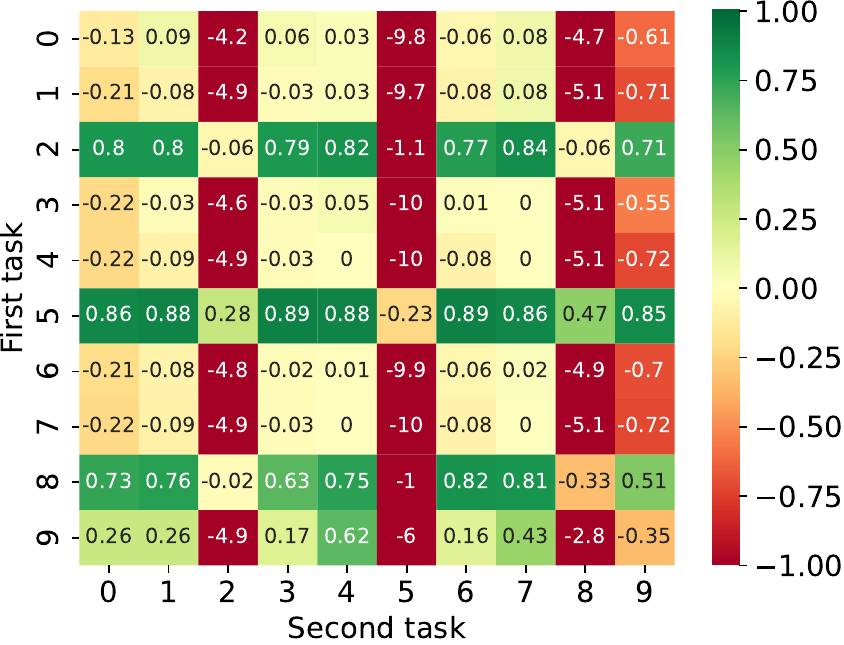}
         \caption{Meta-World}
         \label{fig:tm-meta-world}
     \end{subfigure} 
     %\\ \vspace{5mm}
     \hfill
     \begin{subfigure}[b]{0.47\textwidth}
         \centering
         \includegraphics[width=\textwidth]{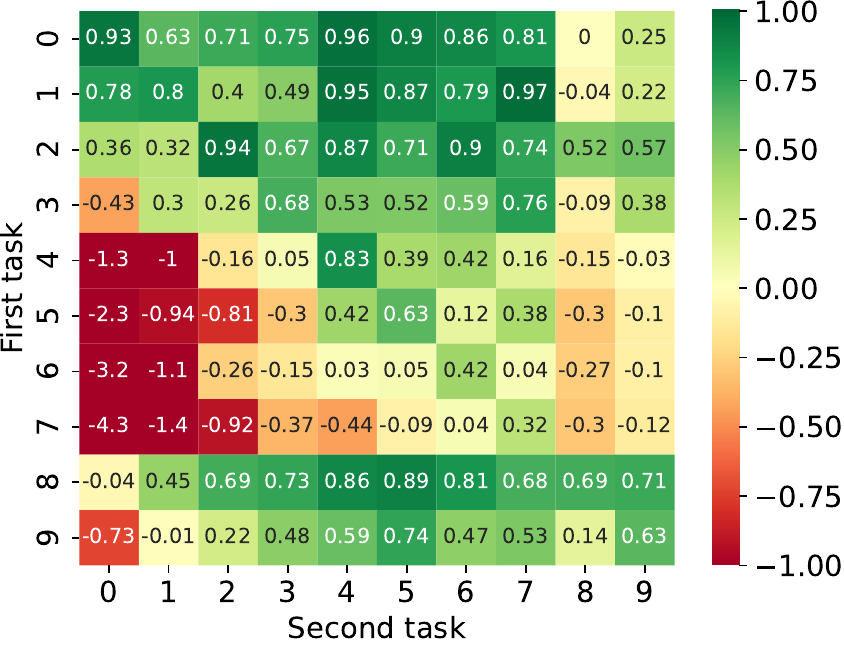}
         \caption{SpaceInvaders}
         \label{fig:tm-spaceinvaders}
     \end{subfigure}
     \\ \vspace{5mm}
     \begin{subfigure}[b]{0.47\textwidth}
         \centering
         \includegraphics[width=\textwidth]{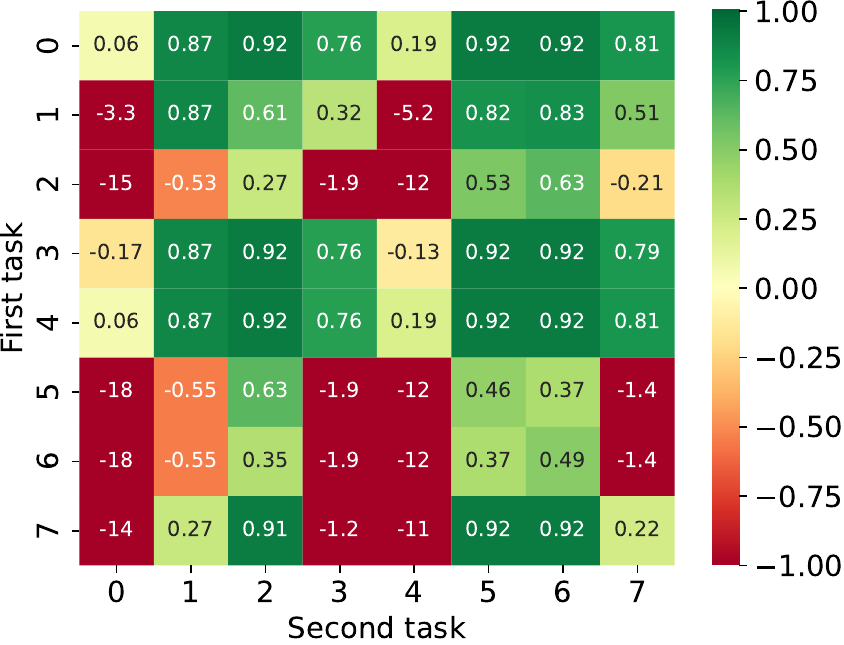}
         \caption{Freeway}
         \label{fig:tm-freeway}
     \end{subfigure}
     \caption{Forward transfer matrices for all sequences. Each element in the matrices is computed as the average forward transfer of training a model from scratch in the first task (Y-axis) and fine-tuning it in the second (X-axis). Results aggregate values from 3 different random seeds. Note that Figure~\ref{fig:tm-meta-world} is a $10\times10$ matrix and not $20\times20$, corresponding with the 10 different tasks that comprise the sequence, as the remaining 10 are repetitions of these.}\label{fig:transf-mats}
\end{figure}

Figure~\ref{fig:transf-mats} presents the forward transfer (FTr) matrices used to compute the RT of each sequence, described in Equation~\eqref{eq:rt} from Section~\ref{sec:metrics} and provided in Table~\ref{table:main-results}. 

Starting from the matrix corresponding to the Meta-World sequence in Figure~\ref{fig:tm-meta-world}, we see that most of the forward transfer values are negative, including the RT (reported in Table~\ref{table:main-results}), indicating considerable interference between the tasks that comprise the sequence. 
However, few tasks that greatly transfer to others, indicated by the mostly green rows when the first task is 2, 5, 8, or 9.
Contrarily, looking at the columns of the matrix, we see that when the second task is 2, 5, or 8, the transfer to these tasks from practically any other task is negative, implying that other tasks provide almost no information to solve the mentioned ones.
Finally, note that the diagonal of the matrix has no especially positive transfer values. 
Even though this might appear to be counterintuitive (as fine-tuning an agent trained from scratch in the same task might result in high FTr), resetting the replay buffer in SAC and learning from instances collected from a mostly uniform policy in the first steps of training (the output head is re-initialized at the beginning of each task) has been shown to cause this effect, as described by \citet{wołczyk2021continual}.

Regarding SpaceInvaders, in Figure~\ref{fig:tm-spaceinvaders}, we observe considerably higher transfer values compared to the previous sequence. In fact,
SpaceInvaders is the task with the highest RT, as shown in Table~\ref{table:main-results}.
Looking at the transfer matrix, we see that the rows corresponding to tasks 0 and 8 have especially higher transfer values, showing that these tasks have a high transfer to the other ones. 
Moreover, in the area of negative values in the center leftmost part of the matrix, we see that tasks of the middle of the sequence (4-7) transfer poorly to the first tasks of the sequence (0-2).   
Note that, in this case, the optimization algorithm is PPO, which has no replay buffer, and thus, the values of the diagonal are all positive and have a considerably high value, although not all are close to 1.
We believe that this might be caused by the stochasticity introduced when re-initializing the last layer of the actor and the critic network every time the task changes.

Figure~\ref{fig:tm-freeway} provides the FTr matrix of the last sequence, Freeway. Likewise SpaceInvaders, we see that, in general, the transfer values in this sequence are mostly positive, however, this sequence shows particularly low values when the transfer is negative. The latter corresponds to the values in the column corresponding to task 0, where the only tasks from which this one benefits are task 4 and itself.
Concerning the diagonal of the matrix, we observe the same behavior as in the previous task. 

\section{Implementation Details} \label{sec:implementation-details}

To make the presented experiments as fair and reproducible as possible, our implementations are based on those from \citet{huang2022cleanrl}, which provides high-quality implementations of RL algorithms. 
Specifically, we maintain modifications to the original SAC and PPO implementations minimal, only changing the definition of the agent to hold the architecture described in Section~\ref{sec:model} and the methods from Section~\ref{sec:experimental-setup}. Note that both SAC and PPO require an actor-critic setting, where the actor is the model from which actions are sampled, while the critic is trained to predict the sum of future rewards given a state. Following common practice in the literature \citep{wolczyk2022disentangling} the CRL methods are only applied to the network of the actor while the critic is restarted at the beginning of each task. All methods share the same common hyperparameters in every task sequence.

In the Meta-World sequence, where SAC is used to optimize all methods, the complete list of hyperparameters is provided in Table~\ref{table:hyperparams-sac}. Following the settings by \citet{yu2020meta}, as with the rest of the hyperparameters for this sequence, the actor and critic networks have been defined as a two-layer MLP, followed by two separate output heads of a single layer, for the mean and the logarithm of the standard deviation of the normal distribution.

Table~\ref{table:hyperparams-ppo} shows the hyperparameters shared by every method in the case of the SpaceInvaders and Freeway sequences under the PPO algorithm. In this case, the same encoder architecture from Appendix~\ref{sec:appendix-encoder} is used for every method to extract low-dimensional feature vectors from image observations. 
Finally, unless otherwise specified, two single-layer output heads are used to generate the logits of the categorical distribution over the action space (actor) and to compute the value function (critic). 
Note that these hyperparameters are set according to the ones employed in the reference implementations from \citet{huang2022cleanrl}.  

%Tables~\ref{table:hyperparams-sac} and~\ref{table:hyperparams-ppo} refer to the hyperparameters shared across all methods in the corresponding task sequences, and the method-specific details are provided in Appendix~\ref{sec:appedix-method-details}.

\begin{table}[ht]
\caption{Hyperparameters shared by all methods in the Meta-World task sequence under the SAC algorithm.}
\label{table:hyperparams-sac}
\vskip 0.15in
\begin{center}
\begin{small}
\begin{sc}
\begin{tabular}{llc}\toprule
& Description & Value \\
\midrule
% Normal
\multirow{7}{*}{Common} & Optimizer & \textit{Adam} \\
& Adam's $\beta_1$ and $\beta_2$  & (0.9, 0.999) \\
& Discount rate ($\gamma$)    & 0.99 \\
& Max Std. & $e^2$ \\ 
& Min Std. & $e^{-20}$ \\ 
& Activation Function & \textit{ReLU} \\ 
& Hidden Dimension ($d_\textit{model}$) & 256 \\ 
\midrule
% SAC specific
\multirow{10}{*}{SAC Specific} & Batch size & 128 \\
& Buffer size & $10^6$ \\
& Target Smoothing Coef. ($\tau$) & 0.005 \\
& Entropy Regularization Coef. ($\alpha$) & 0.2 \\
& Auto. Tuning of $\alpha$ & Yes \\
& Policy Update Freq. & 2 \\
& Target Net. Update. Freq & 1 \\
& Noise Clip & 0.5 \\
& Number of Random Actions & $10^4$ \\
& Timestep to Start Learning & $5\times10^3$ \\
\midrule
% Networks
\multirow{4}{*}{Networks} & Target Net. Layers & 3 \\ 
& Critic Net. Layers & 3 \\ 
& Actor's Learning Rate & $10^{-3}$ \\ 
& Q Networks's Learning Rate & $10^{-3}$ \\ 
\bottomrule
\end{tabular}
\end{sc}
\end{small}
\end{center}
\vskip -0.1in
\end{table}

\begin{table}[ht]
\caption{Hyperparameters shared by all methods in the SpaceInvaders and Freeway task sequences under the PPO algorithm.}
\label{table:hyperparams-ppo}
\vskip 0.15in
\begin{center}
\begin{small}
\begin{sc}
\begin{tabular}{llc}\toprule
    & Parameter                   & Value \\ 
    \midrule
    %%%%%%%%%%%%%%% Common %%%%%%%%%%%%%%
    \multirow{7}{*}{Common} & Optimizer & Adam  \\
    & AdamW's $\beta_1$ and $\beta_2$  & (0.9, 0.999) \\
    & Max. gradient norm          & 0.5 \\
    & Discount rate ($\gamma$)    & 0.99 \\
    & Activation Function         & \textit{ReLU} \\
    & Hidden Dimension ($d_\textit{model}$) & 512 \\ 
    & Learning rate               & $2.5\cdot10^{-4}$ \\
    \midrule
    \multirow{13}{*}{PPO Specific} & PPO Value function coef.    & 0.5 \\
    & GAE $\lambda$               & 0.95 \\
    & Num. parallel environments  & 8 \\
    & Batch size                  & 1024 \\
    & Mini-batch size             & 256 \\
    & Num. mini-batches           & 4 \\
    & Update epochs               & 4 \\
    & PPO Clipping coefficient    & 0.2 \\
    & PPO Entropy coefficient     & 0.01 \\
    & Learn. Rate Annealing       & Yes \\
    & Clip Value Loss             & Yes \\
    & Normalize Advantage         & Yes \\
    & Num. Steps per Rollout      & 128 \\
\bottomrule
\end{tabular}
\end{sc}
\end{small}
\end{center}
\vskip -0.1in
\end{table}

\subsection{Encoder for Visual Control Tasks}\label{sec:appendix-encoder}

As mentioned in Section~\ref{sec:experimental-setup}, the ALE environments (SpaceInvaders and Freeway) define an observation space of $210\times160$ pixels, requiring an encoder network (see Section~\ref{sec:encoder}) to process the images.  
Specifically, we define the same Convolutional Neural Network (CNN) architecture for all methods. Following the implementation from \citet{huang2022cleanrl}, the CNN is composed of three convolutional layers with 32, 64, and 64 channels, with filter sizes being 8, 4, and 3 respectively. The last layer is a dense layer with an output dimension of 512. The rest of the hyperparameters of the encoder network are the ones provided in Table~\ref{table:hyperparams-ppo}.

% %% ====> Hyperparameters
% The hyperparameters and their corresponding values are provided in Table~\ref{table:hyperparams}. 
% For the sake of brevity, the table has been divided into three groups. First, in the \textit{Common} group, the hyperparameters shared across all experiments are given, followed by the ones for the continuous control tasks, and finally the hyperparameters corresponding to the Arcade Learning Environment. 
% The values were set according to the default implementations from \textit{CleanRL}. 
% The hyperparameters of the agents learned from scratch (baseline) and CompoNet have been the same. 
% %% ====> Actor & Critic (+ layer init.)
% Regarding the actor and critic models, separate NNs have been considered, with no learnable shared parameters. Note that the presented architecture has only been applied to the actor and that the critic has been a feed-forward MLP. 
% The weights of the actor and critic models were initialized using orthogonal initialization~\citep{saxe2013exact}, except the linear projections of the attention heads of CompoNet, which employed the uniform Xavier initialization~\citep{glorot2010understanding}. The initial value of the biases was set to zero.   

\subsection{Method Specific Details} \label{sec:appedix-method-details}

The following lines describe all of the considered CRL methods in the experimentation from Section~\ref{sec:experiments}, also providing method-specific implementation details. 

% \paragraph{Baseline} This method consists of training an NN that starts from randomly initialized parameters in every task. Note that this is not a CRL method per se, although it has been used as a baseline for the metrics described in Section~\ref{sec:metrics}. In fact, we believe that training from scratch is a valid approach for CRL, although we should expect CRL methods to at least match its performance or outperform it. 
\paragraph{Baseline.} This method involves training an NN initialized with random parameters for each task. Although not a CRL method \textit{per se}, we believe that training from scratch is a valid approach for CRL, however, we expect CRL-specific  methods to at least match its performance. Moreover, this method serves as the baseline for the forward transfer metric presented in Equation~\eqref{eq:ft} from Section~\ref{sec:metrics}.

\paragraph{FT-1 and FT-N.} FT-1 consists of continuously fine-tuning the same NN from the first to the last task of a sequence. Although this approach usually achieves great results in forward transfer, the final performance is usually relatively low and suffers from considerable forgetting~\citep{wolczyk2022disentangling,wołczyk2021continual,parisi2018role}. This is mainly driven by the fact that FT-1 does not protect the parameters learned in previous tasks when learning new ones, overwriting relevant parameters for other tasks and causing catastrophic forgetting (see Appendix~\ref{sec:further-results}).
Alternatively, FT-N follows the same procedure as the latter method but saves the model trained in each of the $N$ tasks to alleviate the aforementioned issues at the sacrifice of \modi{increasing memory cost}. Following the implementation by \citet{wołczyk2021continual}, the output heads have been re-initialized at the beginning of every task, fine-tuning the preceding layers of the networks.  

\paragraph{Progressive Neural Networks.} ProgressiveNet proposed in \citet{rusu2016progressive} is one of the best-known growing NN approaches \citep{rusu2017sim,parisi2019continual,nips2019compacting,hadsell2020embracing,khetarpal2022continual,gaya2023building}. Every time the task changes ProgressiveNet instantiates a new network, adding lateral connections between the current network and the ones learned in previous tasks (whose parameters are frozen, similarly to CompoNet). Specifically, the input of every layer of the new network will not only consist of the output of the preceding layer but also include the outputs of the layers from the networks learned in previous tasks. Although ProgressiveNet has shown great performance and significant forward transfer abilities across tasks of multiple domains \citep{rusu2016progressive,nips2019compacting,gaya2023building}, the number of parameters required by the approach grows quadratically with respect to the number of tasks, as does its computational cost of inference. 

% \modi{The ProgressiveNet architecture is used in all the layers of the network except the output layer (i.e., the \textit{head})}. 
\modi{Note that the ProgressiveNet architecture consitutes the whole actor network except the final output layer (i.e., the \textit{head}).}
After the end of each task, the last layer is saved and a new layer (with random initial parameters) is instantiated for the new task.     

\paragraph{PackNet.} Proposed by \citet{mallya2018packnet}, this method retains the parameters learned in several tasks in the same network while avoiding catastrophic forgetting. Specifically, once a task is learned, the method prunes the network selecting the parameters relevant to solve the current task, and retrains it to preserve performance. Then, the selected parameters are kept frozen (immutable) for the rest of the future tasks. In each task, the number of iterations to retrain the model is set to $20\%$ of the timestep budget of each task ($200K$ steps, being $\Delta=1M$).

Note that this method requires knowledge of the total number of tasks beforehand, as the number of parameters to be kept for each task ($p/N$, where $p$ is the total number of parameters, and $N$ the number of tasks in the sequence) has to be known in \modi{advance}. This requirement is one of the main issues of PackNet in the context of CRL, as continually learning agents should be able to learn in a variable or an unknown number of tasks \citep{hadsell2020embracing}. Moreover, note that the number of trainable parameters decreases with every task, as $p/N$ parameters are frozen every time a task ends. In fact, after the $N$-th task, no trainable parameters are left in the network.

As with ProgressiveNet, and following the procedure by \citet{wołczyk2021continual}, we apply PackNet to all layers except the last one, which is saved at the end of each task, and re-initialized before starting a new one.

\paragraph{CompoNet.} Regarding the specific implementation details for CompoNet, we have applied the presented architecture to the actor of the SAC and PPO algorithms, using a separate network for the critic.
In the case of SAC, the internal policy of CompoNet has been defined with the parameters in Table~\ref{table:hyperparams-sac}. The output of the module defines the mean vector of the normal distribution employed by the agent, while a separate network has been used for the logarithm of the standard deviation. 
In the image-based task sequences, SpaceInvaders and Freeway, we use a single encoder per module as described in Section~\ref{sec:encoder}, where the encoder of a new module is initialized with the weights of the previous one. Moreover, the output of the encoder, $\bm{h_s}$, served both for the CompoNet module (the actor), and as the input to the value function, defined as a single fully connected layer. 
Finally, the internal policy employs a fully connected network with two layers following the hyperparameters in Table~\ref{table:hyperparams-ppo}.

\subsection{Hardware Resources}

The experimentation has been conducted on two cluster nodes, one containing eight RTX3090 GPUs, an Intel Xeon Silver 4210R CPU, and 345GB of RAM, while the second comprises eight Nvidia A5000 GPUs with an AMD EPYC 7252 CPU and 377GB of RAM. Execution times for the SpaceInvaders and Freeway sequences were approximately 1.5 hours per task, and approximately 3 hours in the case of Meta-World, times mostly varied depending on the method.

\section{Further Experimental Results}\label{sec:further-results}

In this section, we present additional experimental results that complement observations discussed in the main body of the paper, providing a more comprehensive understanding of the performance and behavior of the considered methods.

\subsection{Success Rate Curves}\label{sec:success-rate-curves}

\begin{figure}[H]
    \centering
    \includegraphics[width=0.9\textwidth]{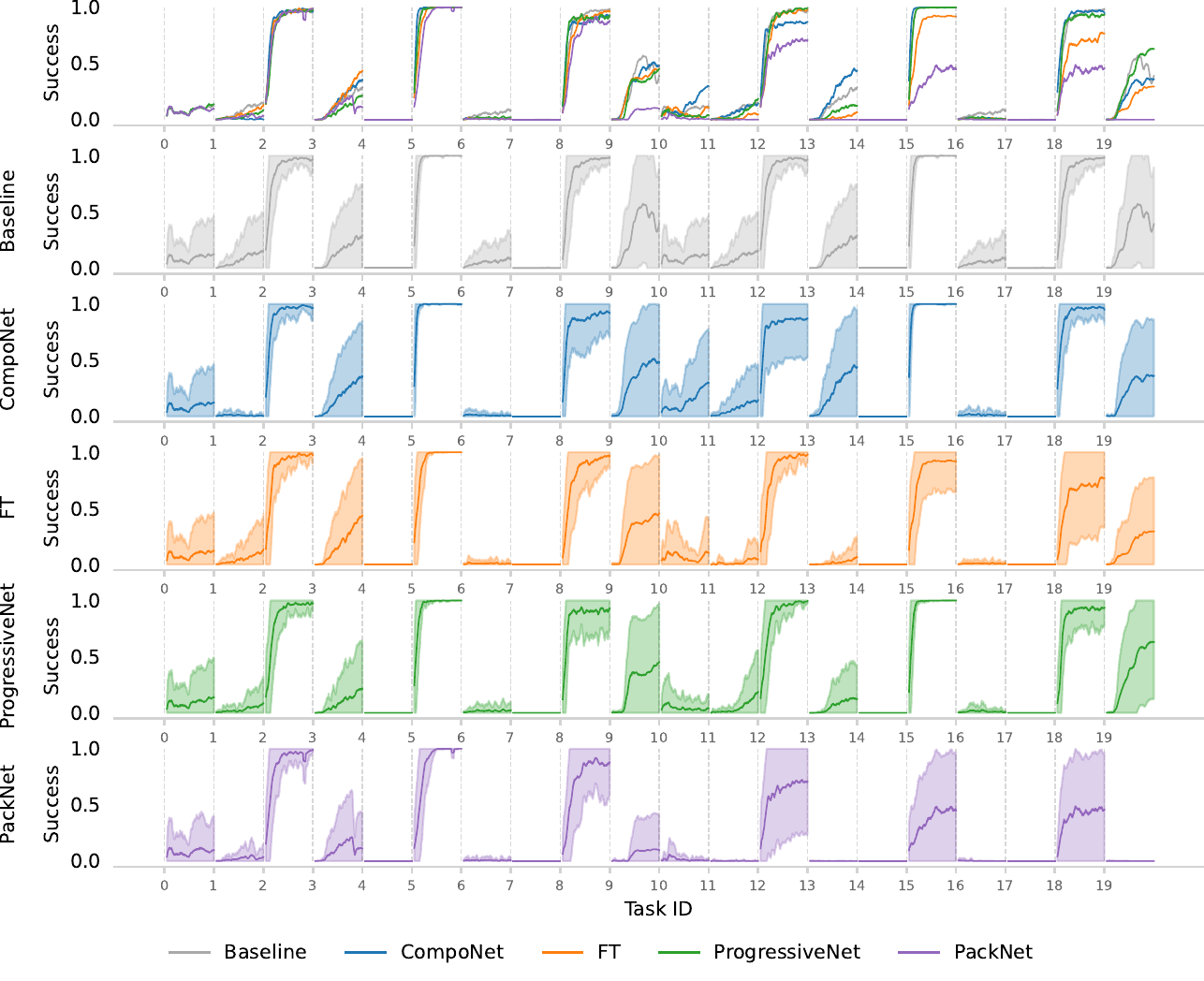}
    \caption{Success curves of all methods in the Meta-World sequence. Results from 10 seeds are aggregated.}
    \label{fig:success-curves-metaworld}
\end{figure}

\begin{figure}[H]
    \centering
    \includegraphics[width=\textwidth]{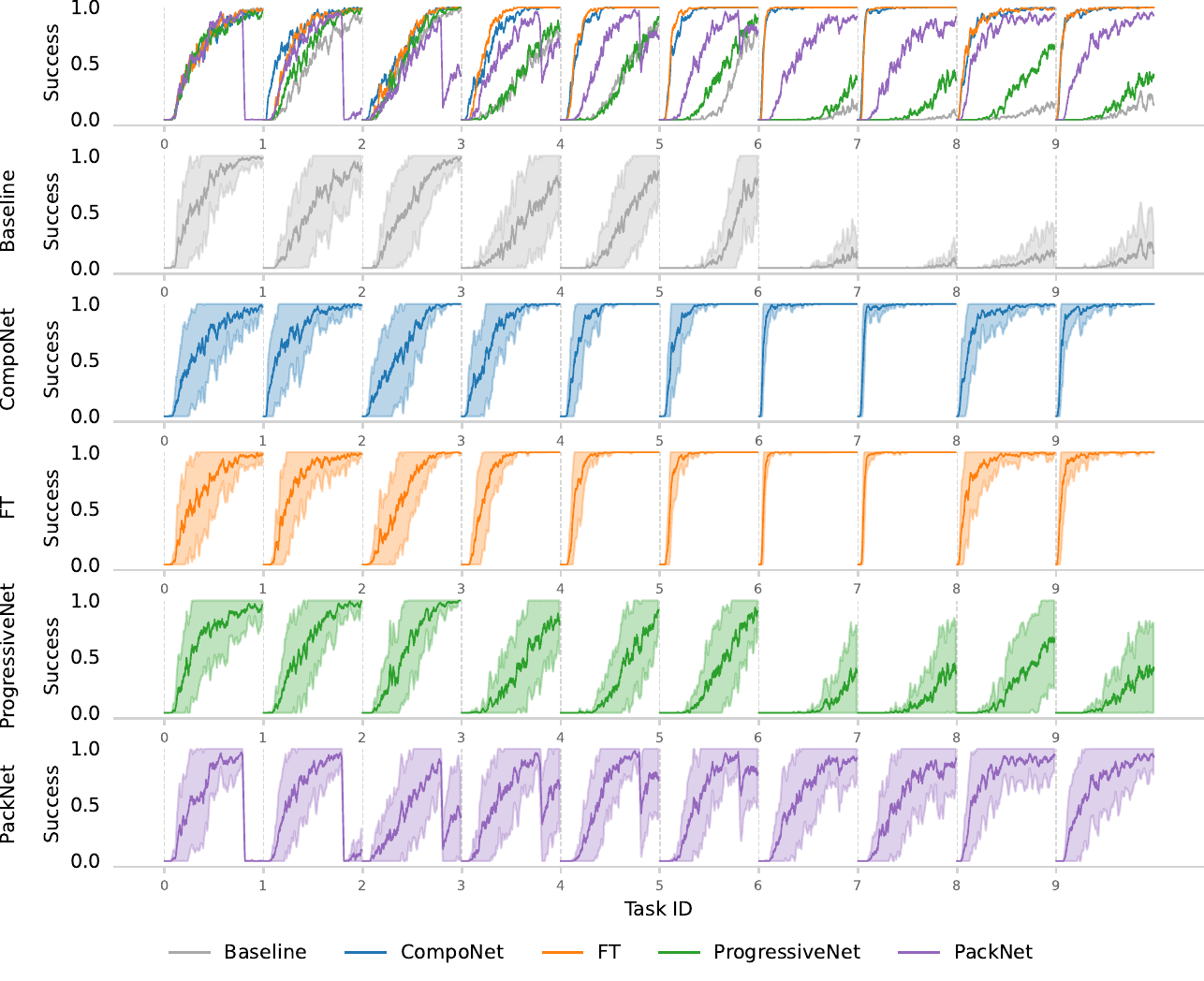}
    \caption{Success curves of the different methods in the SpaceInvaders sequence. Results from 10 seeds are aggregated.}
    \label{fig:success-curves-space-invaders}
\end{figure}

\begin{figure}[H]
    \centering
    \includegraphics[width=\textwidth]{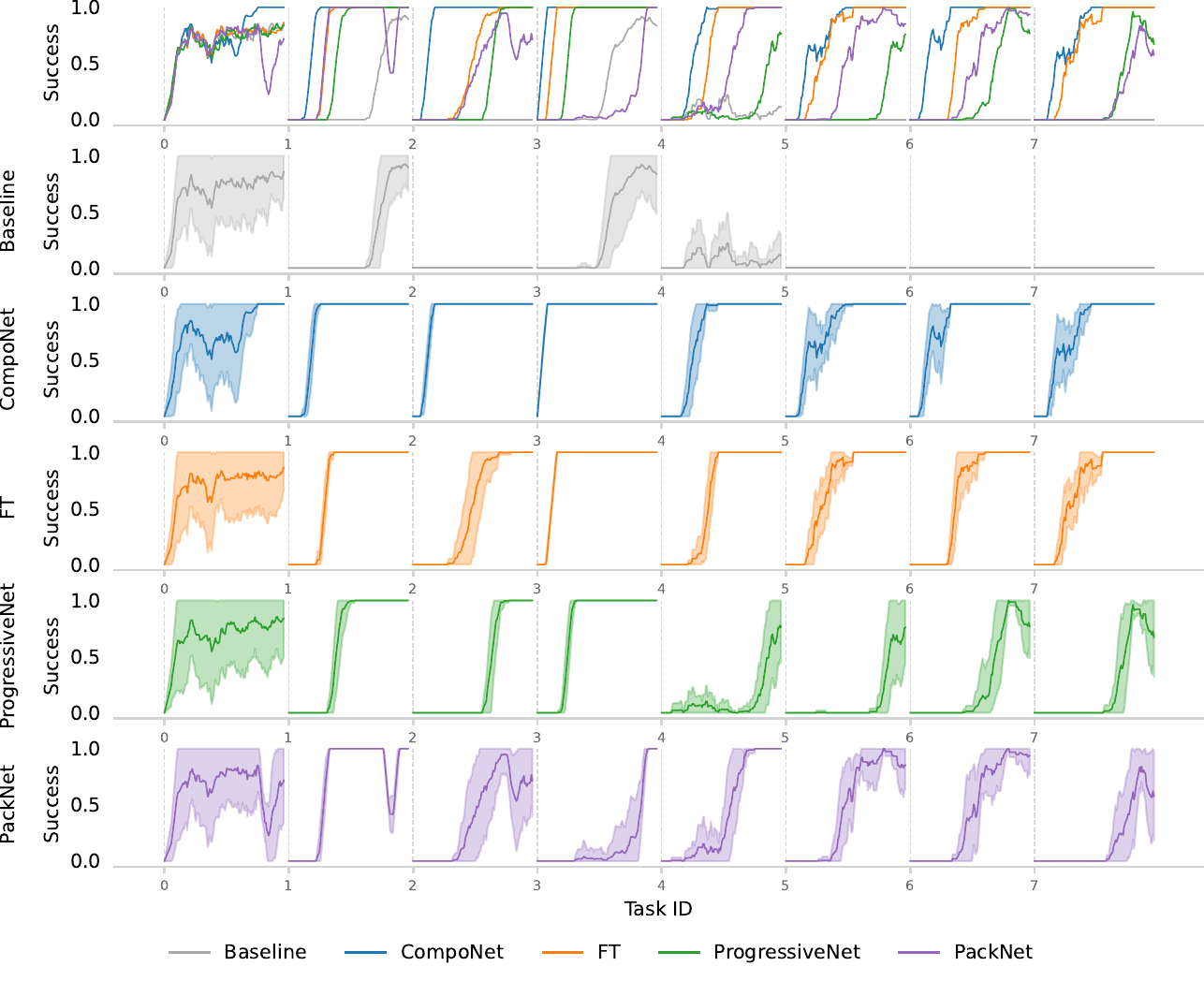}
    \caption{Success curves of the different methods in the Freeway sequence. Results from 10 seeds are aggregated.}
    \label{fig:success-curves-freeway}
\end{figure}

The success rate curves of the methods in the Meta-World, SpaceInvaders, and Freeway sequences are provided in Figures~\ref{fig:success-curves-metaworld}, \ref{fig:success-curves-space-invaders}, and~\ref{fig:success-curves-freeway} respectively. \modi{The} figures show the results from which the CRL relevant metrics from Section~\ref{sec:metrics} have been derived to obtain the values presented in Table~\ref{table:main-results} of the main body of the paper. Each figure aggregates the results of 10 random seeds per method in every task.  The first row illustrates the average curves of all methods, while the rest of the rows depict each method separately including a contour with the standard deviation. Note that the FT-1 and FT-N methods share the same success rate curves, referred to as FT in the figures. \modi{The latter is driven by the fact that both methods share the exact same learning method and hyperparameters, only differing in how they handle forgetting (not measured with this metric).}  

Note that in Figure~\ref{fig:success-curves-metaworld}, corresponding to Meta-World, in tasks 4, 7, 14, and 17 (corresponding to \verb|stick-pull-v2| and \verb|shelf-place|, as tasks 14 and 17 are repetitions of these), none of the methods can obtain a consistent positive success rate. 
Although we employ the hyperparameters in \citet{yu2020meta} for Meta-World, we believe that these results might be caused by the differences in the implementations of the SAC algorithm, as our implementation is based on the one from \citet{huang2022cleanrl}. Nevertheless, we think that considering a few considerably difficult tasks is of special interest in the context of CRL: methods should be robust to any type of scenario, and should not assume that consistent positive success will be achieved for every task.  

Regarding the curves of the SpaceInvaders sequence in Figure~\ref{fig:success-curves-space-invaders}, we observe that every method obtains some positive success rate in each task. 
In the curves of the baseline method, we see that from task 6 and onwards, tasks are especially hard to approach with no information from previous ones. In fact, the rest of the methods, which consider information from preceding tasks, achieve substantially higher success rate values.
Also note that in the first tasks of the sequence (mainly tasks 0 and 1), PackNet reaches high success values, falling to very low values at the end of these tasks. This behavior is caused by the heavy pruning of the network in the first tasks of the sequence, where a high percentage of the parameters are pruned to consolidate the most important parameters for the task at hand. For a description of the method refer to Appendix~\ref{sec:appedix-method-details}.  

Figure~\ref{fig:success-curves-freeway} provides the success rate curves of the last sequence, Freeway. From the curves corresponding to the baseline, we identify that tasks 2, 4, 5, and 7 are of special difficulty within this sequence. This might be caused by the fact that Freeway is a scenario of very sparse reward, as the agent obtains a positive reward only when it reaches its objective, otherwise, the reward is zero. 
Therefore, the mentioned tasks might be particularly hard scenarios for a random initial policy to obtain positive feedback from which to optimize the policy.  
Refer to Appendix~\ref{sec:appendix-freeway} for a detailed description of Freeway and its tasks. 
Concerning PackNet, although the success rate drops when pruning the network (mostly noticeable in the first three tasks of the sequence), the retraining phase effectively updates the network to the same success values obtained before pruning.

%
% Success curves
%
%\newpage
%\newpage
%\newpage

\subsection{Forgetting}\label{sec:forgetting-appendix}

In addition to the metrics described in Section~\ref{sec:metrics}, forgetting is also commonly analyzed in the CRL literature. Following the notation from Section~\ref{sec:metrics}, we define the forgetting in the $i$-th task as, 
\begin{equation}
    \text{F}_i = p_i(i\cdot\Delta) - p_i(T)
\end{equation}
Intuitively, forgetting is the difference between the performance of the model at the end of a task, and the performance in the same task after the end of the whole sequence. 
Therefore, if a model has successfully retained the knowledge obtained in a task, it should achieve near-zero forgetting values. Contrarily, positive forgetting values correspond to scenarios where the model forgets previously learned information. Moreover, when forgetting is negative, the model has been able to acquire information in later tasks that help to solve the previous one. This phenomenon is called \textit{backward knowledge transfer}. 

Under the settings assumed in this work (task boundaries and identifiers are known to the agents, see Section~\ref{sec:preliminaries}), the only methods that can theoretically achieve backward knowledge transfer are FT-1 and the baseline, which precisely suffer from positive forgetting values as shown in the following lines. 
The rest of the methods store the parameters learned in each task, effectively avoiding forgetting (see Appendix~\ref{sec:appedix-method-details} for further details).
Forgetting values for the affected methods are given in Table~\ref{table:all-forgs}, aggregating the results of 10 random seeds per method in every task. As shown, both methods suffer from considerable forgetting, and no backward transfer case is present.

% Note that in the case where task identifiers are known to the agent (as it is in this paper, see the assumptions from Section~\ref{sec:preliminaries}), CompoNet naturally avoids forgetting: if an already solved task is revisited, the policy module that was learned in that task could be used.
% See Appendix~\ref{sec:appedix-method-details} for the description and particular considerations made for all of the considered methods.   

\begin{table}[ht]
\centering
\caption{Forgetting of the baseline and FT-1 methods in all of the considered task sequences. Note that the rest of the methods are omitted as their forgetting is zero under the assumptions from Section~\ref{sec:preliminaries}. Results from 10 random seeds are aggregated.}
\label{table:all-forgs}
%%%%%%%%%%%%%%%%%%%%%%%%%%%%%%%%%%%%%%%
\begin{subtable}{\textwidth}
\centering
\caption{Meta-World}\label{table:forg-meta-world}
\scalebox{0.85}{
\begin{small}
\begin{sc}
\begin{tabular}{lcccccccccc|c}
\toprule 
\multicolumn{1}{l}{Method} & Task 0 & Task 1 & Task 2 & Task 3 & Task 4 & Task 5 & Task 6 & Task 7 & Task 8 & Task 9 & Avg. \\ \midrule
% Baseline & $0.09\pm0.00$ & $0.19\pm0.00$ & $0.90\pm0.00$ & $0.23\pm0.00$ & $0.00\pm0.00$ & $0.97\pm0.00$ & $0.03\pm0.00$ & $0.00\pm0.00$ & $1.00\pm0.00$ & $0.00\pm0.00$ & $0.34\pm0.41$\\
% FT-1 & $0.19\pm0.00$ & $0.12\pm0.00$ & $1.00\pm0.00$ & $0.38\pm0.00$ & $0.00\pm0.00$ & $0.95\pm0.00$ & $0.00\pm0.00$ & $0.00\pm0.00$ & $0.95\pm0.00$ & $0.17\pm0.00$ & $0.38\pm0.39$\\
Baseline & 0.02\scalebox{0.7}{$\pm 0.30$} & 0.15\scalebox{0.7}{$\pm 0.00$} & 0.89\scalebox{0.7}{$\pm 0.27$} & 0.29\scalebox{0.7}{$\pm 0.00$} & 0.00\scalebox{0.7}{$\pm 0.00$} & 0.97\scalebox{0.7}{$\pm 0.17$} & 0.10\scalebox{0.7}{$\pm 0.00$} & 0.00\scalebox{0.7}{$\pm 0.00$} & 0.98\scalebox{0.7}{$\pm 0.00$} & 0.01\scalebox{0.7}{$\pm 0.49$} & 0.34\scalebox{0.7}{$\pm 0.46$} \\
FT-1 & 0.12\scalebox{0.7}{$\pm 0.00$} & 0.19\scalebox{0.7}{$\pm 0.00$} & 0.99\scalebox{0.7}{$\pm 0.00$} & 0.45\scalebox{0.7}{$\pm 0.00$} & 0.00\scalebox{0.7}{$\pm 0.00$} & 0.95\scalebox{0.7}{$\pm 0.22$} & 0.02\scalebox{0.7}{$\pm 0.00$} & 0.00\scalebox{0.7}{$\pm 0.00$} & 0.88\scalebox{0.7}{$\pm 0.10$} & 0.17\scalebox{0.7}{$\pm 0.45$} & 0.38\scalebox{0.7}{$\pm 0.42$}\\
\bottomrule
\end{tabular}
\end{sc}
\end{small}
}
\end{subtable} \\ \vspace{5mm}
%%%%%%%%%%%%%%%%%%%%%%%%%%%%%%%%%%%%%%%
\begin{subtable}{\textwidth}
\centering
\caption{SpaceInvaders}\label{table:forg-space-invaders}
\scalebox{0.85}{
\begin{small}
\begin{sc}
\begin{tabular}{lcccccccccc|c}
\toprule 
\multicolumn{1}{l}{Method} & Task 0 & Task 1 & Task 2 & Task 3 & Task 4 & Task 5 & Task 6 & Task 7 & Task 8 & Task 9 & Avg. \\ \midrule
% Baseline & $0.66\pm0.47$ & $0.85\pm0.34$ & $0.81\pm0.39$ & $0.85\pm0.36$ & $0.99\pm0.10$ & $0.99\pm0.10$ & $1.00\pm0.00$ & $1.00\pm0.00$ & $0.91\pm0.26$ & $0.74\pm0.44$ & $0.88\pm0.32$\\
% FT-1 & $0.58\pm0.49$ & $0.57\pm0.49$ & $0.19\pm0.39$ & $0.32\pm0.47$ & $0.56\pm0.50$ & $0.61\pm0.49$ & $0.60\pm0.49$ & $0.67\pm0.47$ & $0.67\pm0.47$ & $0.70\pm0.46$ & $0.36\pm0.48$ \\
Baseline & 0.66\scalebox{0.7}{$\pm 0.47$} & 0.85\scalebox{0.7}{$\pm 0.34$} & 0.81\scalebox{0.7}{$\pm 0.39$} & 0.85\scalebox{0.7}{$\pm 0.36$} & 0.99\scalebox{0.7}{$\pm 0.10$} & 0.99\scalebox{0.7}{$\pm 0.10$} & 1.00\scalebox{0.7}{$\pm 0.00$} & 1.00\scalebox{0.7}{$\pm 0.00$} & 0.91\scalebox{0.7}{$\pm 0.26$} & 0.74\scalebox{0.7}{$\pm 0.44$} & 0.88\scalebox{0.7}{$\pm 0.32$}\\
FT-1 & 0.58\scalebox{0.7}{$\pm 0.49$} & 0.57\scalebox{0.7}{$\pm 0.49$} & 0.19\scalebox{0.7}{$\pm 0.39$} & 0.32\scalebox{0.7}{$\pm 0.47$} & 0.56\scalebox{0.7}{$\pm 0.50$} & 0.61\scalebox{0.7}{$\pm 0.49$} & 0.60\scalebox{0.7}{$\pm 0.49$} & 0.67\scalebox{0.7}{$\pm 0.47$} & 0.67\scalebox{0.7}{$\pm 0.47$} & 0.70\scalebox{0.7}{$\pm 0.46$} & 0.36\scalebox{0.7}{$\pm 0.48$} \\
\bottomrule
\end{tabular}
\end{sc}
\end{small}
}
\end{subtable} \\ \vspace{5mm}
%%%%%%%%%%%%%%%%%%%%%%%%%%%%%%%%%%%%%%%
\begin{subtable}{\textwidth}
\centering
\caption{Freeway}\label{table:forg-freeway}
\scalebox{0.85}{
\begin{small}
\begin{sc}
\begin{tabular}{lcccccccc|c}
\toprule 
\multicolumn{1}{l}{Method} & Task 0 & Task 1 & Task 2 & Task 3 & Task 4 & Task 5 & Task 6 & Task 7 & Avg. \\ \midrule 
% Baseline & $0.02\pm0.44$ & $0.87\pm0.00$ & $0.49\pm0.35$ & $1.00\pm0.00$ & $0.51\pm0.44$ & $0.80\pm0.24$ & $0.70\pm0.24$ & $0.60\pm0.43$ & $0.62\pm0.42$ \\
% FT-1 & $0.49\pm0.45$ & $0.75\pm0.33$ & $0.62\pm0.11$ & $0.88\pm0.33$ & $0.54\pm0.42$ & $0.69\pm0.37$ & $0.62\pm0.34$ & $0.72\pm0.33$ & $0.58\pm0.43$ \\
Baseline & 0.02\scalebox{0.7}{$\pm 0.44$} & 0.87\scalebox{0.7}{$\pm 0.00$} & 0.49\scalebox{0.7}{$\pm 0.35$} & 1.00\scalebox{0.7}{$\pm 0.00$} & 0.51\scalebox{0.7}{$\pm 0.44$} & 0.80\scalebox{0.7}{$\pm 0.24$} & 0.70\scalebox{0.7}{$\pm 0.24$} & 0.60\scalebox{0.7}{$\pm 0.43$} & 0.62\scalebox{0.7}{$\pm 0.42$} \\
FT-1 & 0.49\scalebox{0.7}{$\pm 0.45$} & 0.75\scalebox{0.7}{$\pm 0.33$} & 0.62\scalebox{0.7}{$\pm 0.11$} & 0.88\scalebox{0.7}{$\pm 0.33$} & 0.54\scalebox{0.7}{$\pm 0.42$} & 0.69\scalebox{0.7}{$\pm 0.37$} & 0.62\scalebox{0.7}{$\pm 0.34$} & 0.72\scalebox{0.7}{$\pm 0.33$} & 0.58\scalebox{0.7}{$\pm 0.43$} \\
\bottomrule
\end{tabular}
\end{sc}
\end{small}
}
\end{subtable} \\ \vspace{5mm}
\end{table}

\section{Ablation Study}\label{sec:ablations}

In Section~\ref{fig:arch-val} CompoNet is validated on several edge cases, showing that it fulfills the expected behavior in the scenarios that guided its development (introduced at the beginning of Section~\ref{sec:model}). 
This appendix aims to provide additional evidence on the fact that the three blocks that build the self-composing policy module (see Figure~\ref{fig:compo-unit}) cooperate to achieve the expected behavior. 

\subsection{Output Attention Head}\label{sec:ablation-out-head}

This appendix analyzes the contribution of the output attention head to the overall performance and forward transfer abilities showed by CompoNet in the experiments in Section~\ref{sec:experiments}. 

\begin{figure}[ht]
     \centering
     \begin{subfigure}{\textwidth}
         \centering
         \includegraphics[width=\textwidth]{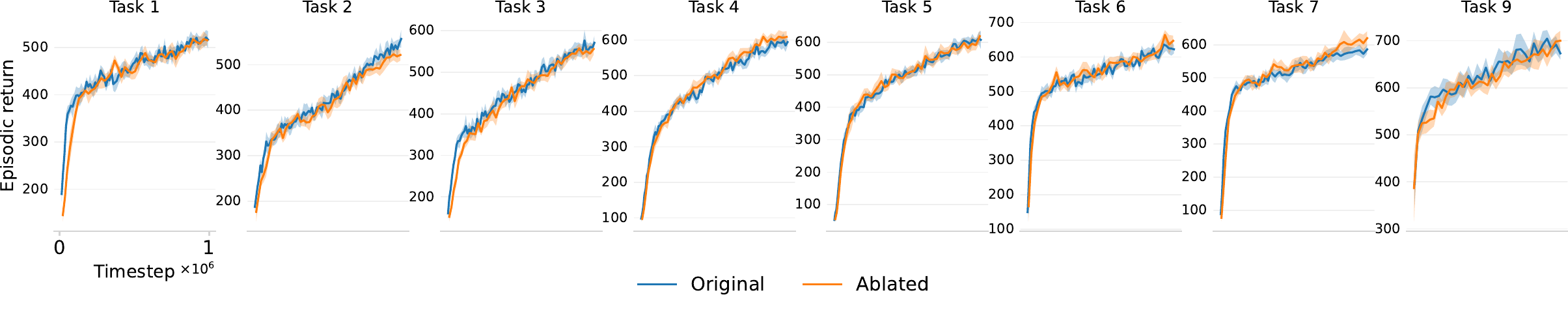}
         \caption{SpaceInvaders}
         \label{fig:out-head-ablation-space-invaders}
     \end{subfigure}\\

     \vspace{5mm}
     
     \begin{subfigure}{\textwidth}
         \centering
         \includegraphics[width=\textwidth]{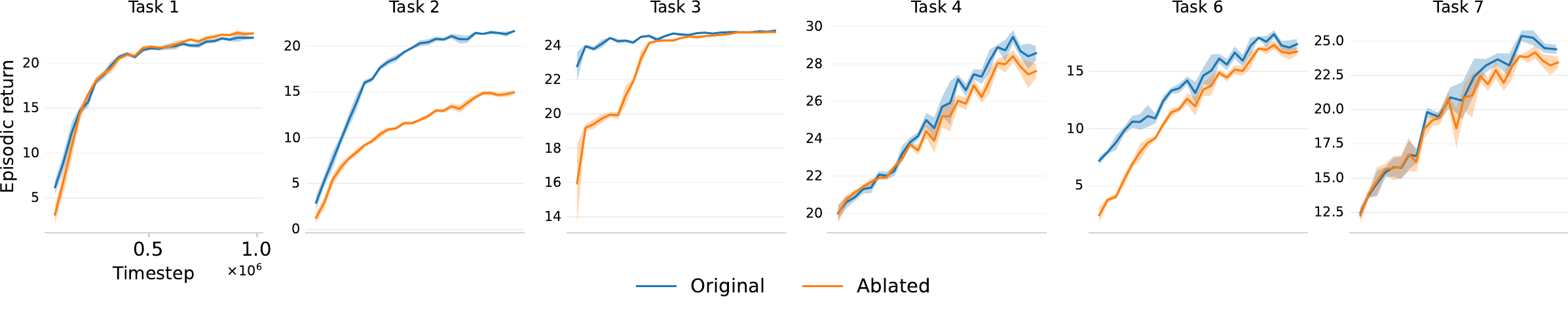}
         \caption{Freeway}
         \label{fig:out-head-ablation-freeway}
     \end{subfigure} 
     \caption{Episodic return curves for the SpaceInvaders and Freeway sequences. Blue lines correspond to the original CompoNet architecture presented in Section~\ref{sec:model}, while orange lines represent the same architecture with no output attention head (referred to as \textit{Ablated}). Each curve is computed from the average result of 5 seeds and contours indicate standard error.}
     \label{fig:out-head-ablation}
\end{figure}

Results are shown in Figure~\ref{fig:out-head-ablation}, where the performance of the original CompoNet design and an ablated version without the output attention head are compared in the SpaceInvaders and Freeway sequences, Figures~\ref{fig:out-head-ablation-space-invaders} and~\ref{fig:out-head-ablation-freeway} respectively.
% SpaceInvaders
The performance of the original and the ablated versions do not considerably differ in the SpaceInvaders sequence, except for slightly better results for the original CompoNet design in the first timesteps of some tasks, see Figure~\ref{fig:out-head-ablation-space-invaders}.
% Freeway
These results greatly contrast with the ones obtained in the Freeway sequence (see Figure~\ref{fig:out-head-ablation-freeway}), where the ablated version shows considerably lower performance in every task except the first.

Due to the design of CompoNet (see Figure~\ref{fig:compo-unit} and Section~\ref{sec:self-compo-pol}), the output attention head should be of special interest in tasks where reusing previous modules is the best-performing strategy. 
In this sense, we believe that results from Figure~\ref{fig:out-head-ablation} are mainly affected by the fact that the Freeway environment is a scenario highly sparse reward (see the description of the sequence in Appendix~\ref{sec:appendix-freeway}). This fact makes the reuse of previously learned policies a strategy of special interest. This hypothesis is supported by the low performance obtained by the baseline in a great part of the tasks depicted in Figure~\ref{fig:success-curves-freeway}.

\subsection{Input Attention Head}\label{sec:ablation-in-head}

In this appendix, we evaluate the influence of the input attention head on the performance and forward transfer capabilities exhibited by CompoNet in the experiments presented in Section~\ref{sec:experiments}.

For this purpose, we have designed an experimental setting where CompoNet should employ the input attention head to gather information to solve the current task, while this information can not be directly used to solve the task via imitation using the output attention head.
Specifically, we have trained a CompoNet agent in the fifth task of the SpaceInvaders sequence, where the module operating in the current task has access to five previous modules: four non-informative (sample their output from a uniform Dirichlet distribution), and one trained to solve the current task. This setting is equivalent to the one presented in the leftmost plots of Figure~\ref{fig:arch-val}, although in this case the probability vectors outputted by all previous modules have been shifted one element to the left.\footnote{By shifting one element to the left we mean that each output vector $\bm{v}=(v_1, v_2, \ldots, v_{n-1}, v_n)$ is modified as $\bm{v}' =(v_2, v_3, \ldots, v_n, v_1)$.} This implies that the probability originally assigned to each action is assigned to another, nullifying the strategy of directly using previously learned policies as the output of the current module.

\begin{figure}[ht]
    \centering
    \includegraphics[width=\textwidth]{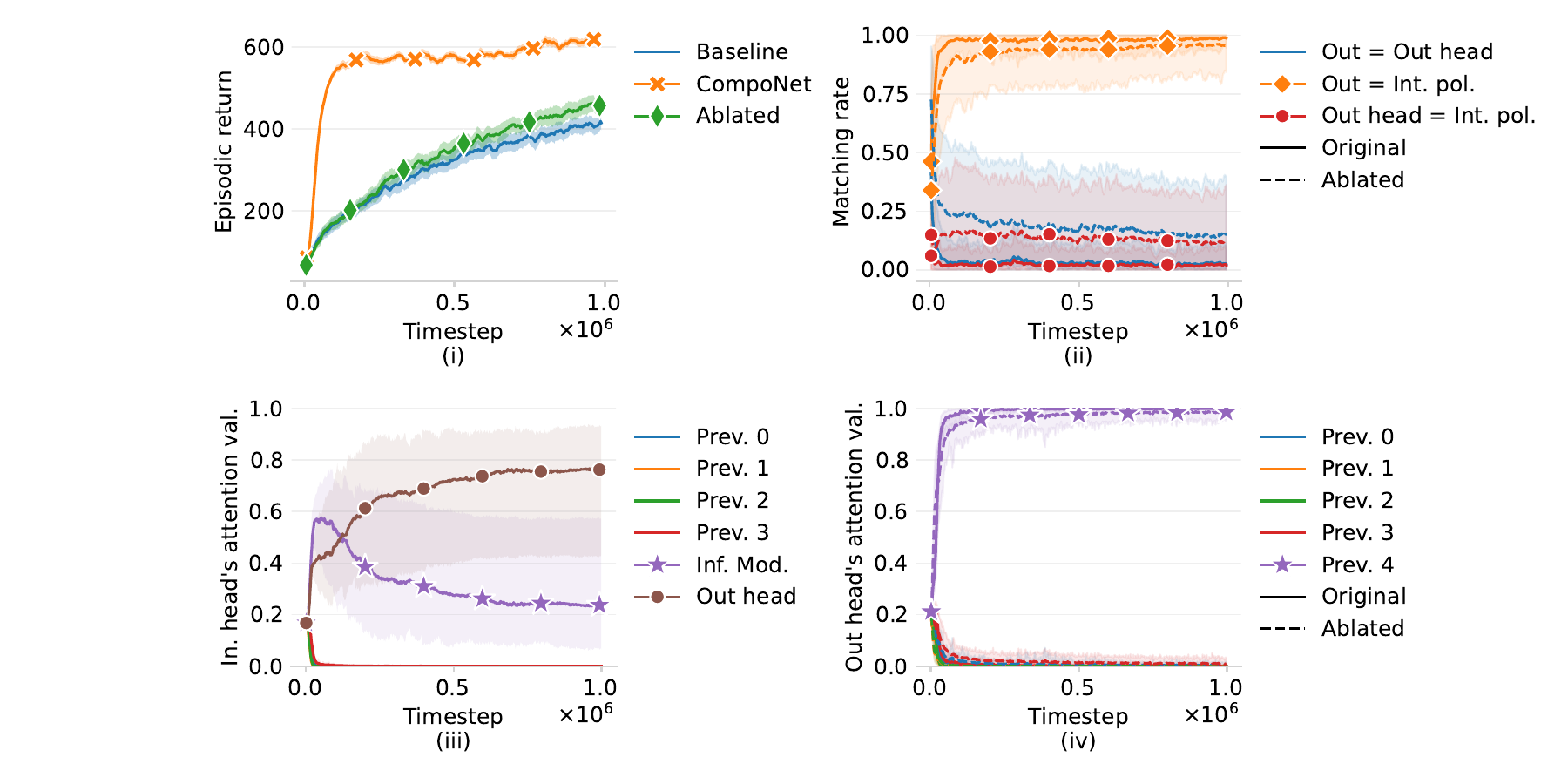}
    \caption{Comparison of the original CompoNet design, an ablated version of CompoNet without the input attention head, and the baseline. The methods have been trained to solve the fifth task of the SpaceInvaders sequence using the hyperparameters from Appendix~\ref{sec:implementation-details}, where the results of 10 seeds have been aggregated. In the case of the CompoNet models, five previous modules have been provided: four non-informative, and one trained to solve the current task (called \textit{Inf. Mod.} and marked with stars). However, the output vectors of these modules have been shifted one element to the left, making the direct usage of \textit{Inf. Mod.} an unviable approach although still maintaining useful information to solve the current task. 
    \modi{In (i) the performance of the methods is compared in terms of episodic return; (ii) depicts the rate in which the outputs of different parts of the models match; finally, (iii) and (iv) respectively illustrate the evolution of the attention values for the input and output heads of the module being trained.}
    Results show that the original version of CompoNet successfully employs the input attention head to extract information for the internal policy to solve the task at hand, greatly outperforming the ablated version and the agent trained from scratch.}
    \label{fig:ablation-input-head}
\end{figure}

Results are presented in Figure~\ref{fig:ablation-input-head}, where the module trained to solve the current task is named \textit{Inf. Mod.} (marked with stars). The figure aggregates the results of 10 random seeds for the baseline, the original CompoNet architecture, and an ablated version of CompoNet that has no input attention head.\footnote{$\bm{h_s}$ is the only input of the internal policy in the ablated CompoNet model.} 
Observing the episodic return curves in Figure~\ref{sec:ablation-in-head}.i, we can see that the original CompoNet architecture successfully employs the information at hand to reach substantially higher episodic returns faster. 
Moreover, in Figure~\ref{sec:ablation-in-head}.iii, we see that CompoNet attends to the module with information on the current task (\textit{Inf. Mod.}) and to the output attention head, which in turn only attends to \textit{Inf. Mod.} (see Figure~\ref{sec:ablation-in-head}.iv). Finally, although in both versions of CompoNet, the output attention head learns to attend \textit{Inf. Mod.}, the final output of the model is defined by the internal policy (see Figure~\ref{sec:ablation-in-head}.ii), especially in the case of the non-ablated version, where the final output of the model completely matches the output of the internal policy.

In summary, the ablated version of CompoNet is capable of learning a policy from scratch with minimal interference, as shown by the similar episodic return curve to the baseline. However, leveraging the input attention head, CompoNet successfully extracts information from previous policies to substantially enhance the training in the current task. 

% \newpage

\section{Performance of the attention heads in extremely long task sequences}\label{sec:att-scalability}

\modi{
In this appendix, we focus on another aspect of scalability when dealing with extremely long task sequences beyond the computational costs (already analyzed in detail in Appendixes~\ref{sec:cost-inference} and \ref{sec:memory-cost-appendix}).
Specifically, we analyze the ability of the proposed method to efficiently capture information from previous modules when the number of these is extremely large.
For this purpose, we incorporate a previous module trained to solve the current task and multiple non-informative ones that sample their outputs from a uniform Dirichlet distribution. 
This setting resembles the experiment presented in the first part of Section~\ref{sec:archi-val} and is illustrated in the leftmost plots of Figure~\ref{fig:arch-val}, but in this case, the number of non-informative previous modules is increased and is substantially larger.}

\begin{figure}[ht]
    \centering
    \includegraphics[width=0.5\textwidth]{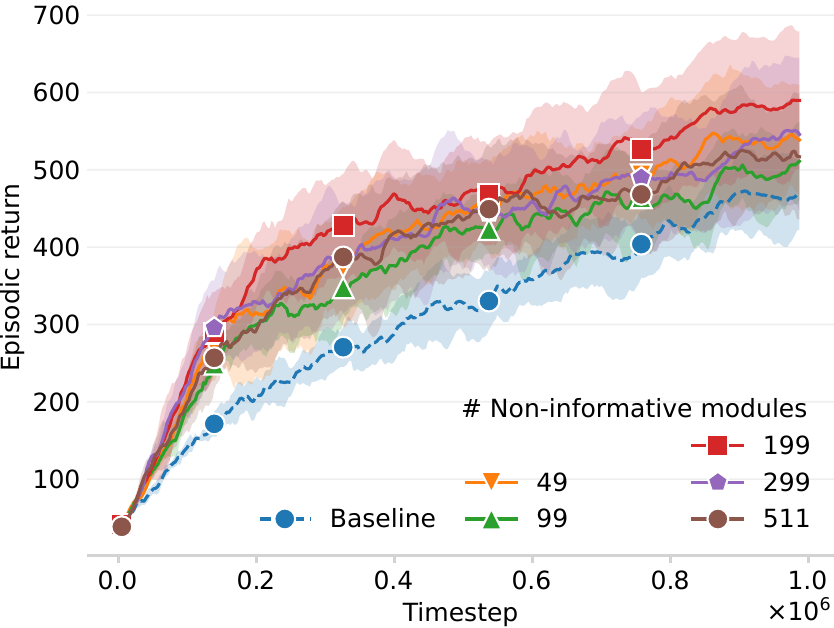}
    \caption{\modi{Scalability of the attention heads to retrieve useful information from previous modules in extremely long task sequences. The figure compares the episodic return curve of the baseline method (trained from scratch) to CompoNet with an increasing number of non-informative previous modules and a single module that is trained to solve the current task. Specifically, the task at hand is the sixth task of the SpaceInvaders sequence. Lines and contours respectively represent the mean and standard deviation of 5 runs with different random seeds. Note that the total number of previous modules for each curve is the number of non-informative modules, shown in the legend, plus one, corresponding to the informative (i.e., trained) module.}}
    \label{fig:att-scalability}
\end{figure}

\modi{
Results are shown in Figure~\ref{fig:att-scalability}. The first observation is that the episodic return curves of the CompoNet agents grow faster and to greater values than the agent trained from scratch to solve the task at hand (referred to as the baseline method). This demonstrates that the attention heads capture relevant information from the informative previous module despite the much larger number of non-informative ones, showing the robustness of CompoNet to a large number of modules.
}

\modi{Moreover, when comparing the curves corresponding to CompoNet, we observe that there is no clear correspondence with the number of non-informative previous modules and performance. For instance, the curve corresponding to 511 non-informative previous modules is almost consistently above the one regarding 99 of such modules. This suggests that the number of non-informative modules could be further increased without a substantial loss in performance. Note that, despite these results, we expect the number of non-informative previous modules will affect the performance of CompoNet when the number of previous modules is increased by orders of magnitude to the ones tested in this experiment. However, note that the 512 previous modules tested in this experiment would correspond to a sequence of 512 previous tasks, which is much larger than the usual sequences in the literature, where 20 tasks are considered rather a large sequence, and 100 tasks have been considered at most to the best of our knowledge \cite{wołczyk2021continual}.}

%%%%%%%%%%%%%%%%%%%%%%%%%%%%%%%%%%%%%%%%%%%%%%%%%%%%%%%%%%%%%%%%%%%%%%%%%%%%%%%
%%%%%%%%%%%%%%%%%%%%%%%%%%%%%%%%%%%%%%%%%%%%%%%%%%%%%%%%%%%%%%%%%%%%%%%%%%%%%%%

\end{document}